\pdfoutput=1
\documentclass[11pt]{article}

\usepackage[final]{acl}

\usepackage{times}
\usepackage{latexsym}

\usepackage[T1]{fontenc}
\usepackage[utf8]{inputenc}

\usepackage{microtype}

\usepackage{inconsolata}

\usepackage{graphicx}

\usepackage{latexsym}
\usepackage{tabularx} % \usepackage{arydshln}
\usepackage{booktabs}
\usepackage{multirow}
\usepackage{subfigure}
\usepackage{amsmath}
\usepackage{amssymb}
\usepackage{array}
\usepackage{bm}
\usepackage{xparse}
\usepackage{mathtools}
\usepackage{pifont}
\usepackage{enumitem}
\usepackage{tcolorbox}
\tcbuselibrary{breakable}
\usepackage{listings}
 
\usepackage{arydshln}
\usepackage{hyperref}

\title{Evaluating Personalized Tool-Augmented LLMs from the Perspectives of \\ Personalization and Proactivity}

\author{
 \textbf{Yupu Hao\textsuperscript{1,2}},
 \textbf{Pengfei Cao\textsuperscript{1,2}},
 \textbf{Zhuoran Jin\textsuperscript{1,2}},
 \textbf{Huanxuan Liao\textsuperscript{1,2}},
\\
 \textbf{Yubo Chen\textsuperscript{1,2}},
 \textbf{Kang Liu\textsuperscript{1,2}},
 \textbf{Jun Zhao\textsuperscript{1,2}},
\\
 \textsuperscript{1}The Key Laboratory of Cognition and Decision Intelligence for Complex Systems, \\Institute of Automation, Chinese Academy of Sciences, Beijing, China\\
 \textsuperscript{2}School of Artificial Intelligence, University of Chinese Academy of Sciences, Beijing, China
\\
\{haoyupu2023, liaohuanxuan2023\}@ia.ac.cn, \\ \{pengfei.cao, zhuoran.jin, yubo.chen, kliu, jzhao\}@nlpr.ia.ac.cn
}

\begin{document}
\maketitle
\begin{abstract}
Personalized tool utilization is essential for aligning large language models (LLMs) with user preference in interaction scenarios with various tools. However, most of the current benchmarks primarily focus on either personalization of text generation or direct tool-utilizing, without considering both. In this work, we introduce a novel benchmark \textbf{ETAPP} for evaluating personalized tool invocation, establishing a sandbox environment, and a comprehensive dataset of 800 testing cases covering diverse user profiles. To improve the accuracy of our evaluation, we propose a key-point-based LLM evaluation method, mitigating biases in the LLM-as-a-judge system by manually annotating key points for each test case and providing them to LLM as the reference. Additionally, we evaluate the excellent LLMs and provide an in-depth analysis. Furthermore, we investigate the impact of different tool-invoking strategies on LLMs' personalization performance and the effects of fine-tuning in our task. The effectiveness of our preference-setting and key-point-based evaluation method is also validated. Our findings offer insights into improving personalized LLM agents. Our Code is available at \url{https://github.com/hypasd-art/ETAPP}.

\end{abstract}

\section{Introduction}
With the advancement of large language model (LLM) capabilities \cite{zhao2024surveylargelanguagemodels}, more researchers are shifting their alignment objectives from targeting the general human population to focusing on specific small groups or individuals \cite{chen2024personapersonalizationsurveyroleplaying, jang2023personalized}. Personalizing the LLMs refers to adjusting the behavior and output of LLMs to match the needs of individual users better. In this context, \citeauthor{li2024personalllmagentsinsights}\shortcite{li2024personalllmagentsinsights} introduce the concept of the \textit{Personal LLM Agents}, which integrates personalized user data and devices to provide user with comprehensive, continuous, and personalized service. To achieve this goal, the Personal LLM Agents must possess two key capabilities: memory \cite{zhang2024surveymemorymechanismlarge} and interaction \cite{10.1145/3704435, mialon2023augmentedlanguagemodelssurvey}. The memory capability allows the model to track and retain both the user's current state and their historical information, enabling more personalized and context-aware interactions. Meanwhile, the interaction capability empowers the model to engage with external systems and devices, utilizing tools to perform tasks and support the user’s needs in real-time \cite{li-etal-2023-api, DBLP:conf/iclr/QinLYZYLLCTQZHT24}.

Currently, Personal LLM Agent evaluations mainly focus on text generation \cite{zhao2025do}, with limited attention to external interactions. Only a few recent studies have begun to assess Personal LLM Agents in interactive scenarios involving tool utilization. For example, \citeauthor{wang2024aipersonalifelongpersonalization}\shortcite{wang2024aipersonalifelongpersonalization} constructs a life-long personal agent framework and evaluates it using LLM. Although their work incorporates tool calling by simulating the external API execution, it does not conduct a specialized evaluation of the interaction process itself. \citeauthor{cai2024largelanguagemodelsempowered}\shortcite{cai2024largelanguagemodelsempowered} evaluates personalized tool usage in the specific domain of online shopping with a limited set of tools. However, these works still fail to answer the following core questions: (1) Compared to normal tool-utilizing, what evaluation metrics should be designed to assess personalized tool-utilizing? (2) How can we effectively evaluate the personalized tool usage capability in more realistic environments, beyond the narrow domain and the simulated scenarios?

\begin{figure*}[t]
  \centering
  \includegraphics[width=0.95\textwidth]{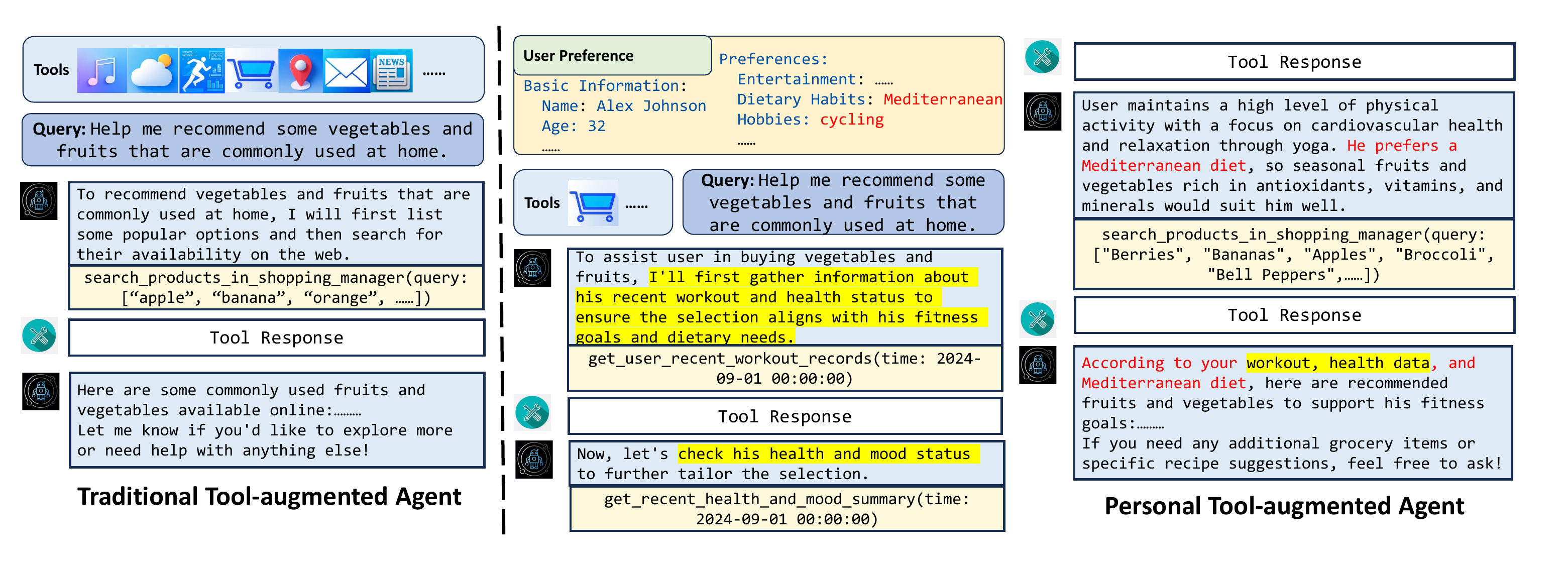} 
  \caption{The difference between a traditional tool-augmented agent and a personal tool-augmented agent. Red font represents output reflecting personalization, while yellow background font represents output reflecting proactivity.
  }
    \label{fig:different_model}
\end{figure*}

To answer the first question, we argue that an ideal personal assistant should not only provide personalized services but also understand the user's intentions and anticipate unspoken needs. It should offer comprehensive support by considering factors beyond immediate instructions, easing the user's burden. From this perspective, Figure~\ref{fig:different_model} illustrates the distinctions between traditional tool-augmented LLMs and personal tool-augmented LLMs. The latter should exhibit two key features: personalization and proactivity. \textbf{Personalization} ensures the model tailors its responses and tool usage based on the user's preferences and needs. For example, when the user asks for food recommendations, the personalized LLM considers that the user prefers a Mediterranean diet (highlighted in red font), in contrast to traditional agents that offer popular options. \textbf{Proactivity} refers to the model anticipating and suggesting actions beyond the user's request to help complete tasks more comprehensively. For instance, the assistant checks the user's recent health status and workout records to further tailor the recommendation (highlighted in yellow background), unlike traditional models, which overlook these factors. This action is not required by the user but effectively enhances the quality of services.

Based on the two metrics mentioned above, we construct a new benchmark called \textbf{E}valuation of \textbf{T}ool-augmented \textbf{A}gent from the \textbf{P}ersonalization and \textbf{P}roactivity Perspective (\textbf{ETAPP}) to evaluate the personalized tool invocation capabilities of large language models. For the tool-invoking framework, we build a simple sandbox to ensure the stability of the testing process, comprising the following key elements: (1) Environment Setup: We develop a tool-invoking system with 33 functional APIs (e.g., \textit{add\_calendar}, \textit{view\_calendar}, \textit{get\_weather}) belonging to 9 categories (e.g., \textit{Calendar}, \textit{Weather}), including software APIs and hardware APIs. To enhance the authenticity and stability of the evaluation, we design a simple sandbox environment where the API responses are not influenced by external environmental changes. (2)	Memory Building: The memory of personal LLM agents includes long-term user preferences and short-term user status. We divide user preferences into two categories: high-level user profiles and low-level tool-utilizing preferences, which help us to capture user needs more precisely. We generate 16 different user profiles based on professional backgrounds, which serve as the foundation for generating diverse evaluation samples. To simulate real usage scenarios, we construct 9 days of interaction history for each necessary tool and user.
(3)	Instruction Construction: We manually label 50 test instructions, and combine them with 16 different user profiles to create a final dataset of 800 testing cases. This data supports our evaluation of personalized tool invocation capabilities from two perspectives: personalization and proactivity.

To address the second question, we use LLM as the evaluating model to score the performance of the tool-invoking system \cite{li2025generationjudgmentopportunitieschallenges, li2024llmsasjudgescomprehensivesurveyllmbased}. To improve the evaluation reliability, we design a key-point-based LLM evaluation method. Each testing instruction is associated with several key points annotated by humans, which serve as standard indicators for task completion. These key points are provided to the evaluation LLM to assist in scoring. The evaluation model first analyzes whether each point is satisfied and provides final analysis and score. Experimental results demonstrate that these key points significantly enhance the evaluation accuracy.

Finally, we evaluate current excellent LLMs with tool invocation capabilities on the entire set and some reasoning models (e.g., o1-mini, DeepSeek-R1) on a testing subset. The results may suggest that LLMs tend to answer the question directly without deeply reasoning why choosing this tool or have insufficient consideration of the personalized tool-utilizing process. Furthermore, we analyze the impact of different tool-invoking methods on the performance and the results point out the importance of the reasoning process in tool-utilizing. Additionally, we manually label a portion of the testing data and fine-tune a 7B model using these data. The results show that fine-tuning improve performance on in-domain instruction data but has limited effect on out-of-domain instruction data.

To summary, the contribution of our work includes:
\begin{itemize}
    \item We develop a new benchmark for personalized tool invocation and establish a stable and consistent sandbox environment for evaluation. Using a dataset of 800 test cases with various user profiles and preferences, we evaluate the model's performances from both personalization and proactivity perspectives.
    \item We propose a key-point-based LLM evaluation method utilizing manually annotated key points to assist the evaluation for each testing data, improving the reliability compared with directly evaluating. 
    \item We specifically evaluate the effectiveness of our preference design and evaluation methods. We analyze the impact of different tool-invoking methods and conduct an in-depth analysis of how fine-tuning affects model performance in different scenarios.
\end{itemize}

\section{Dataset Construction}
\begin{figure}[t]
  \centering
  \includegraphics[width=\columnwidth]{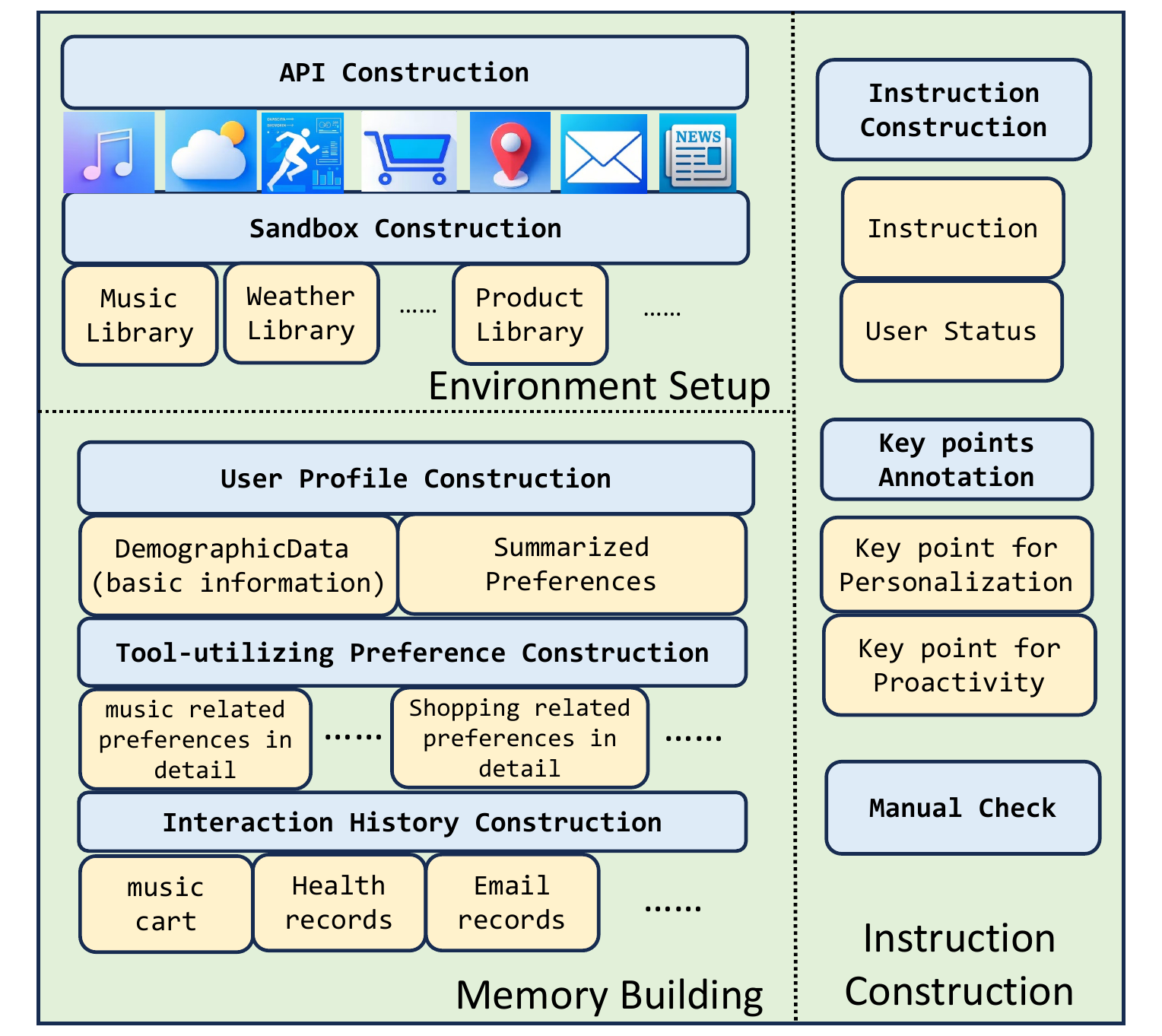} 
  \caption{The process of dataset construction.
  }
    \label{fig:process_construction_dataset}
\end{figure}

As shown in Figure~\ref{fig:process_construction_dataset}, this process includes the following components: Environment Setup, Memory Building, and Instruction Construction.

\subsection{Environment Setup}
To better simulate real-world application scenarios, we manually construct a tool-invoking environment comprising 33 functional APIs (e.g., \textit{add\_calendar}, \textit{get\_weather}) belonging to 9 categories and 2 tool retrieval APIs (\textit{search\_tools} and \textit{get\_tool\_doc}) to search for usable tools. These APIs encompass a wide range of common software APIs (e.g. Calendar and Email) and hardware-dependent APIs (e.g. health monitoring through smart wristbands and smart home control), ensuring a broad range of practical use cases.

Additionally, we develop a simple sandbox environment to ensure the stability and reliability of the testing process. This sandbox isolates the experiment from external environmental variables (e.g., connection error or change of world), thereby preventing any external factors from influencing the API outputs and ensuring the accuracy of the evaluations. Real-world API data is used to generate corresponding outputs (e.g., data from RapidAPI), enhancing the authenticity of the simulated scenarios. For further details, please refer to Appendix~\ref{sec:dataset_construction}.

\subsection{Memory Building}
The core of personalization of LLM lies in effectively utilizing the memory of the user, which can be divided into long-term memory (capturing the user's preferences, habits, and historical behaviors) and short-term memory (reflecting the user's current state like position). % and 

\subsubsection{Long-term and Short-term Memory}
For long-term memory, we propose a novel architecture by subdividing user preferences into two categories to manage the memory more effectively: \textbf{high-level preferences} (User Profile) and \textbf{low-level preferences} (Tool-utilizing Preference). 
The User Profile includes basic information about the user (such as name, age, and occupation) and a summary description of their preferences, while the Tool-utilizing Preference includes a detailed description of the user's preferences for the API in the corresponding category. For example, for APIs related to the music category, we have established a unified music-utilizing preference, which includes attributes such as favorite music, singer, listening habits, etc., providing more detailed guidance for the model to invoke corresponding tool and generate responses. We establish eight types of tool-utilizing preferences for each user (with no preferences designed for the \textit{Weather} category). When a corresponding tool is invoked, the relevant category of preference is input into the model in advance, serving as contextual information to assist the model in generating personalized outputs.

Inputting all preferences at once is neither practical nor efficient due to the vast amount of user data and context window limitations. Traditional methods \cite{zhao2025do}, which summarize preferences from dialogue history, often result in sparse and incomplete coverage, with some preferences not even being reflected in conversations. In contrast, summarizing preferences through API perspective is more comprehensive and accurate.

In detail, we conduct the following memory construction process with the help of LLM. First, the LLM generates 16 different user profiles based on various professions. These profiles are manually verified for diversity and consistency to ensure they cover a wide range of user types. Secondly, based on the above profile, the LLM further generates 8 tool-utilizing preferences for each user, which are then manually verified to ensure relevance and accuracy.

For short-term memory, we define the user's current state for each testing data, including location and current time, reflecting the user's personalized needs more accurately.

\subsubsection{Interaction History Construction}
It is essential to note that some tools rely on the user's interaction history (e.g., retrieving recent exercises or email records). So we need to construct the interaction history for specific APIs. 

To ensure the consistency of the interaction history, we first build a 9-day arrangement for each user and construct an interaction history based on it, covering schedule arrangements, alarms, health status (updated hourly), exercise records, email records, and music collections. The specific steps are in Appendix~\ref{sec:appendix_instruction_history}.

\subsection{Instruction Construction}
\label{sec:instruction_construction}
We manually label 50 unique instructions, combining them with 16 predefined user profiles to create a total of 800 testing instructions. For each instruction, we provid the user status (current time and location of the user). To ensure the accuracy of the interaction history and avoid potential conflicts, we further verify the 9-day interaction history again to ensure every instruction can get an answer. In addition, we manually write the key points for each instruction. We define these key points as the specific requirements that need to be met within our predefined environment to achieve personalization and proactivity. These key points are used for the evaluation in Section~\ref{sec:evaluation_framework}. For statistics of our dataset, please refer to Table~\ref{table:statistics}.

\begin{figure*}[t]
  \centering
  \includegraphics[width=\textwidth]{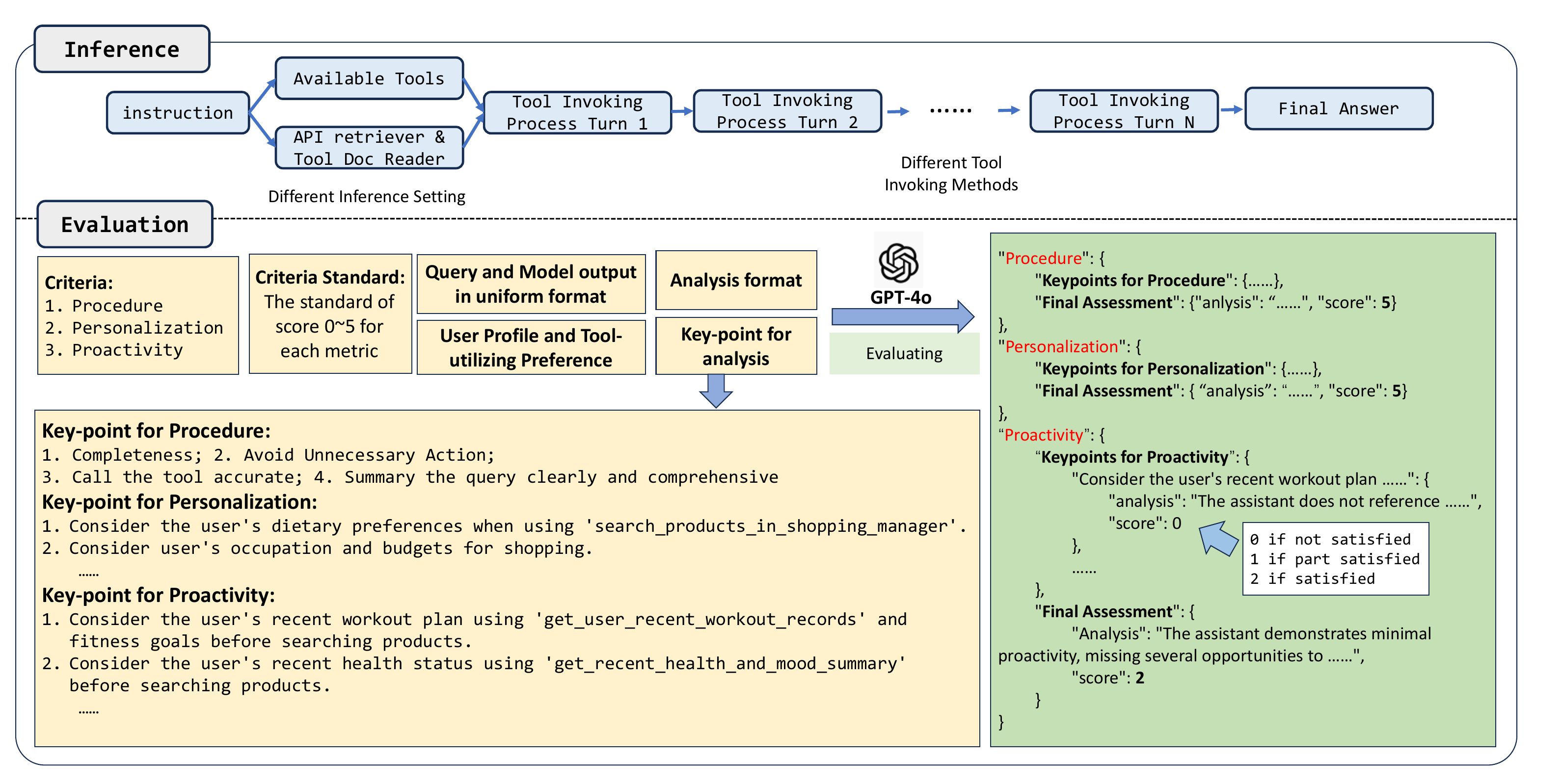} 
  \caption{The process of Inference and Evaluation of our benchmark.
  }
    \label{fig:process}
\end{figure*}
\section{Evaluation}
\subsection{Evaluation Metrics}
The evaluation process employs a large language model for scoring, with criteria based on three dimensions, each rated on a scale of 0 to 5:
(1)	\textbf{Procedure (PRC)}: This metric assesses whether the model provides a complete and accurate final response. The scoring criteria are based on whether the model addresses all key aspects of the user's query and delivers a clear, comprehensive solution without omissions.
(2)	\textbf{Personalization (PSN)}: This metric evaluates whether the model appropriately incorporates the user's preferences and current status into its response, tailoring its solution to the user’s historical data and requirements.
(3)	\textbf{Proactivity (PTV)}: This metric measures whether the model exceeds the user’s explicit instructions by taking additional, meaningful steps to assist the user effectively, proactively identifying user needs and offering extra suggestions or actions, rather than merely reacting to the user’s request.

\subsection{Evaluation Settings}
The evaluation involves two evaluation settings:
(1)	\textbf{Tool-Given}: In this setting, the tools are pre-specified, and the evaluation focuses on the model’s ability to complete the task effectively with known tools.
(2)	\textbf{Tool-Retrieval}: This setting mirrors real-world scenarios, where the model operates within a limited context window, searching for tools with their corresponding utilizing preferences based on task requirements and invoking them to accomplish the tasks.

Based on the data and methods mentioned above, we construct a complete Inference framework as illustrated in the Figure~\ref{fig:process}. The framework consists of the following components: for user $u$ and instructions $Q$, system prompt $I$, available tools $T_Q$, user preferences of the high-level profile $P_h^u$, user tool-utilizing preferences $\{P_t^u|t \in T_Q\}$, user state $C_Q^u$, one-shot example (optional) $E$. These elements are provided as input to the model, which outputs either a tool invocation or a final response. The LLM processes the instruction by interacting with tools, terminating once the final result is obtained or after reaching a predefined maximum step.

In the Tool-Given and Tool-Retrieval setting, the process is formalized as follows respectively:

\begin{equation}
\footnotesize
\begin{aligned}
    A_n&:\text{Tool\_Given} = LLM(I, T_Q^u, P_h^u, \{P_t^u|t \in T_Q\}, \\ &C_Q^u, E, Q, \sum_{i=1}^{n-1}(A_i, O_i))\\
    A_n&:\text{Tool\_Retrieval} = LLM(I, T_Q^u, P_h^u, C_Q^u, E, Q, \\ &\sum_{i=1}^{n-1}(A_i, O_i, \{P_t^u|t \in o_i^j \And a_i^j=\text{tool\_search}\}))
\end{aligned}
\end{equation}

where, an step $i$, $A_i$ is the output of LLM, $a_i^j$ is the j-th tool calling in $A_i$, $O_i$ represents the observation of corresponding tool.

\subsection{Evaluation Framework}
\label{sec:evaluation_framework}
We propose the key-point-based LLM evaluation framework. Due to the non uniqueness of the solution path and the multi-dimensionality and difficulty of preference evaluation, common methods generally use LLMs as evaluators in evaluating tool utilizing ability \cite{DBLP:conf/iclr/QinLYZYLLCTQZHT24} or preference alignment \cite{zhao2025do}. However, we find that directly evaluating LLM performance is difficult because the evaluation model has limited understanding of personalization and proactivity and cannot easily judge whether the output meets the required standards. To improve this, we need to encourage the evaluation model $E\_LLM$ to analyze the model's output through concrete aspects, so we provide the key points written manually to the evaluation LLM.

As shown in Figure~\ref{fig:process}, for each instruction $Q$ and user $u$, the criteria standard of the score from 0 $\sim$ 5 is provided in system prompt $I_E$. The user profile preference and tool-utilizing preferences, and the output of the model is provided to the evaluation model. Besides, the key points for each metric $K$ and the analysis format of the output $F$ are provided. The evaluation model outputs the analysis and score (0 $\sim$ 2) for each key point and gives the final score for the model's performance of instruction. The formula is as follows:

\begin{equation}
\footnotesize
\begin{aligned}
    & \text{Analysis}, \text{Score} = E\_LLM(Q, I_E, P_h^u, \\ &\{P_t^u|t \in T_Q\}, \sum_{i=1}^{N}(A_i, O_i), K, F)
\end{aligned}
\end{equation}

\begin{table*}[t]
\centering
{
    
	\centering
    \resizebox{1\textwidth}{!}{
	\begin{tabular}{l | c | c c c | c  c  c | c } %
        \hline
		% \toprule

        \multirow{2}{*}{\textbf{Model}} 
        & \multirow{2}{*}{\textbf{Method}}
        & \multicolumn{3}{c|}{\textbf{Tool Given}}
        & \multicolumn{3}{c|}{\textbf{Tool Retrieval}}
        & \multirow{2}{*}{\textbf{Average}}
        \\

		&
		& \textbf{PRC}
		& \textbf{PSN}
        & \textbf{PTV}
        & \textbf{PRC}
		& \textbf{PSN}
        & \textbf{PTV}
        &
        \\
		\midrule

        GPT-4o
        & \multirow{5}{*}{\textbf{FC}}
		& \textbf{3.95} 
		& \textbf{3.37} 
        & \textbf{1.61}
        & 2.67
        & 2.19
        & \textbf{1.08}
        & \textbf{2.48}
        
		\\
        % \cline{1-1}
        % \cline{3-9}
        % \midrule
  
		DeepSeek-V3
        &
		& 3.02 
        & 2.78 
        & 1.47
        & 2.57
        & \textbf{2.38}
        & 0.95 
        & 2.20

        \\
        % \cline{1-1}
        % \cline{3-9}
        % \midrule

        Llama-3.1-70B-Instruct
        &
		& 2.31 
        & 1.71 
        & 0.55
        & 1.50
        & 1.06
        & 0.30
        & 1.24

        \\
        % \cline{1-1}
        % \cline{3-9}
        % \midrule

        Qwen2.5-72B-Instruct
        &
		& 3.76 
		& 3.34 
        & 1.5 
        & \textbf{2.72} 
        & 2.31 
        & 1.05
        & 2.45
        
        \\
        % \cline{1-1}
        % \cline{3-9}
        % \midrule

        watt-tool-70B
        &
		& 2.70 
		& 1.97 
        & 0.71
        & 0.84
        & 0.32
        & 0.12
        & 1.11
        
		\\
        \hline
        \hline
        % \midrule
        % \midrule

        GPT-4o
        & \multirow{7}{*}{\textbf{ReAct}}
		& 3.97 
		& 3.61 
        & 1.60
        & 3.70
        & 3.43
        & 1.56
        & 2.98
        
		\\
        % \cline{1-1}
        % \cline{3-9}
        % \midrule
  
		DeepSeek-V3
        &
		& \textbf{4.05} 
        & \textbf{3.78}
        & 1.84
        & \textbf{3.82}
        & \textbf{3.54}
        & \textbf{1.65} 
        & \textbf{3.11}

        \\
        % \cline{1-1}
        % \cline{3-9}
        % \midrule

        o1-preview *
        &
		& 3.67 
		& 3.69 
        & \textbf{1.87}
        & 3.28
        & 3.48
        & 1.60
        & 2.93
        
		\\
        % \cline{1-1}
        % \cline{3-9}
        % \midrule
  
		o1-mini *
        &
		& 3.63 
        & 3.35 
        & 1.61
        & 3.25
        & 3.14
        & 1.35 
        & 2.72

        \\
        % \cline{1-1}
        % \cline{3-9}
        % \midrule

        DeepSeek-R1 *
        &
		& 2.41 
        & 2.06 
        & 1.35
        & 0.93
        & 0.40
        & 0.31
        & 1.24

        \\
        % \cline{1-1}
        % \cline{3-9}
        % \midrule

        DeepSeek-R1-Distill-Qwen-32B *
        &
		& 2.40 
		& 2.57 
        & 1.07 
        & 1.74 
        & 1.66 
        & 0.62
        & 1.68
        
        \\
        % \cline{1-1}
        % \cline{3-9}
        % \midrule

        QwQ-32B-Preview *
        &
		& 1.01 
		& 1.19 
        & 0.53
        & 0.61
        & 0.48
        & 0.18
        & 0.67
        
		\\
        \hline
        % \midrule

		% \bottomrule
	\end{tabular}
}
}
	\caption{The overall results of different LLMs under two settings. * represents testing on a subset of data. The evaluation model we used is GPT-4o.
	}
	\label{table:overall_result}
\end{table*}

\section{Experiments}
\subsection{Baselines}
We adopt two frequently used tool-invoking methods: \textbf{FC (Function Calling)}: After fine-tuning, the model calls the tool through a predefined format; \textbf{ReAct} \cite{yao2023react}: The model combines reasoning with tool invocation, where reasoning guides the model to analyze the current status and decide the appropriate action. The one-shot example is provided to ReAct in this process.

We use the excellent closed-source and open-source models with the function calling ability as baselines in two experimental settings. 
Additionally, we evaluate the reasoning model on a subset of the 100 testing data due to the budgets factors using ReAct method, as they do not support FC. For further details refer to Appendix~\ref{sec:evaluation_details}.

\subsection{Evaluation Results}
The results are shown in Table~\ref{table:overall_result}, we can find that in FC result: (1) In the Tool-Retrieval scenario, the model's performance is significantly lower than in the Tool-Given scenario, indicating that the model struggles more as the length and difficulty of tool-invoking and planning increases; (2) Currently, in FC setting, GPT-4o performs excellently across multiple metrics, while Qwen2.5-72B-Instruct outperforms DeepSeek-V3, demonstrating its strong performance; (3) Even the best-performing model in BFCL \cite{berkeley-function-calling-leaderboard}, Watt-tool-70B, struggles with this tool-invoking task. This could be due to two reasons: a) Not generalizing. Post-training may hinder its ability to fully grasp personalization and proactivity within the tool-planning process; b) Not deep reasoning. During tool learning, the model tends to directly invoke tools based on explicit instructions, without deep reasoning, leading to difficulties in effectively handling Proactivity scenarios without clear instructions. This inspires us to not only enhance the model's ability to use tools but also to think about why invoking this tool.

The reasoning models do not show an advantage over non-reasoning models in this task, particularly the DeepSeek-Distill and QwQ models, which performed poorly. The reasoning models may not integrate the reasoning process with the tool-invoking process, or may not have deep thinking on Personalization and Proactivity in the specific scenario. Some error examples can be found in the Appendix~\ref{sec:appendix_example_error}.

\begin{figure*}[t]
  \centering
  \includegraphics[width=\textwidth]{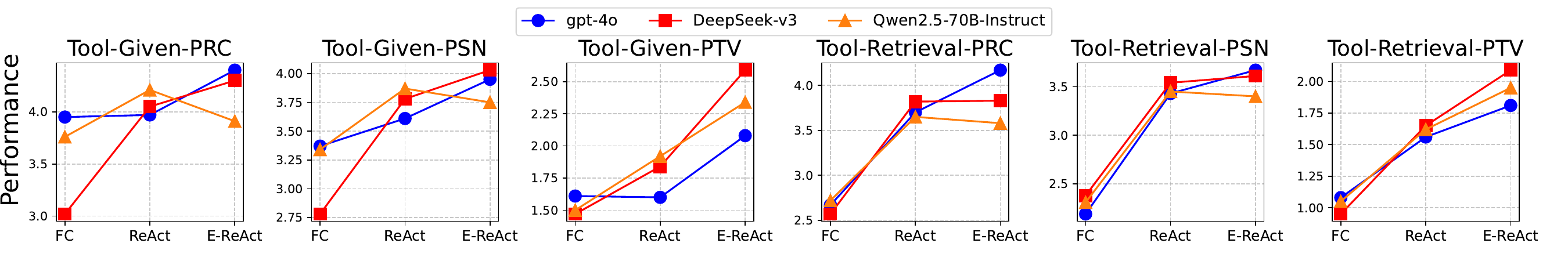} 
  \caption{The performance of different tool-invoking methods.
  }
    \label{fig:different_inference_performance}
\end{figure*}

\subsection{Analysis}
\subsubsection{Tool-invoking Method Analysis}
In order to analyze the impact of different tool-invoking methods on model performance, we adopt the following three approaches: \textbf{FC}, \textbf{ReAct} and \textbf{E-ReAct (Enhanced-ReAct)}. The E-ReAct method extends ReAct by first requiring the model to generate key aspects of personalization, proactivity before invoking tools. Once these key points are addressed, the model proceeds with tool invocation using the ReAct framework to complete the task.

As shown in Figure~\ref{fig:different_inference_performance}, we find that E-ReAct methods can effectively improve the personalization and proactivity capabilities of the model's output. By incorporating a mandatory deep thinking and understanding process before tool invocation, the model better interprets user intentions and provides more proactive services. This suggests that integrating more high-quality reasoning into the tool-learning process may lead to even better performance.

\begin{table}[]
    \centering
    \resizebox{0.9\columnwidth}{!}{
    \begin{tabular}{c|c|c|c|c}
        \midrule
        Setting & Tokens & PRC & PSN & PTV
        \\
        \hline
         All & 3444 &  4.09 & \textbf{3.80}  & 1.62 \\
         \hline
         Needed & \textbf{2393} & \textbf{4.16} & 3.73 & \textbf{1.76}
         \\
         \hline
    \end{tabular}}
    \caption{The results of different preference settings are evaluated on a subset of data. The \textit{Tokens} metric represents the average number of tokens for each instruction calculated by the tokenizer of Llama-3.1-70B-Instruct. And the performance is tested on GPT-4o. 
    }
    
    \label{table:different_preference_setting}
\end{table}
\subsubsection{Preference Setting Analysis}
We divide the preference into the User Profile and Tool-utilizing Preference. Table~\ref{table:different_preference_setting} shows the difference between inputting all preferences into the LLM and our method. We can find that our setting (Needed) reduces the context length and demonstrates competitive results compared to inputting all (All). Notably, we have only included 8 categories of preferences, as the number of categories increases or the preferences become more concrete, the advantage of our setting is expected to become even more prominent.

\subsubsection{Fine-tuning Experiments Analysis}
To validate our idea of the importance of reasoning, we annotate some data in both ReAct and FC format respectively and conduct fine-tuning experiments on Qwen2.5-7B-Instruct using 200 data points. The annotating process is as follows: 
(1)	Generate reference answers using the LLM.
(2)	Perform manual verification and labeling the answers to ensure quality and consistency.
For data in FC format, one set only includes the tool-calling process, while the other also contains the textual thinking before the tool invocation.

The results are in the Table~\ref{table:fine-tuning-result}. We find that: (1) The performance improves significantly for ID instructions and OOD users. However, with the phenomenon improvement for OOD instructions being significantly lower than for ID instructions, this indicates that for instructions within the same scenario, once the model learns the tool invocation process, it can generalize and solve problems even when user preferences differ. However, when a new scenario emerges, the model struggles to reason effectively about how to solve the problem, resulting in limited improvement. (2) Compared to ReAct and FC models, the improvement for ``wo r'' is not significant, demonstrating that a certain level of reasoning process is necessary to enhance the model's multi-turn tool invocation capability. (3) Tool-invoking methods like ReAct still show advantages in this model compared to FC.

\begin{table}[t]
\centering
{
   \renewcommand{\arraystretch}{0.9} 
	\centering
    \resizebox{1\columnwidth}{!}{
	\begin{tabular}{l | c c c } %
		\toprule

        \textbf{Model}

		& \textbf{PRC}
		& \textbf{PSN}
        & \textbf{PTV}

        \\
		\midrule

        \multicolumn{4}{c}{\textbf{U(ID)I(OOD)}}
        \\
        \midrule
        \midrule

        Vanilla (ReACT)
		& 2.76 
		& 2.97 
        & 1.35
        
		\\
        % \midrule
  
		\quad FT (ReAct)
		& \textbf{3.47} ($\uparrow$ 25.8\%) 
        & \textbf{3.28} ($\uparrow$ 10.4\%) 
        & \textbf{1.99} ($\uparrow$ 47.4\%)

        \\
        \midrule
        % \midrule

        Vanilla (FC)
		& 3.08 
        & 2.66 
        & 1.07

        \\
        % \midrule

        \quad FT (FC)
		& 3.23 ($\uparrow$ 4.9\%) 
		& 3.00 ($\uparrow$ 12.8\%) 
        & 1.72 ($\uparrow$ 60.7\%) 
        
        \\

        % \midrule

        \quad FT (FC wo r)
		& 2.67 ($\downarrow$ 13.3\%)
		& 2.68 ($\uparrow$ 0.7\%)
        & 1.62 ($\uparrow$ 51.4\%)
       
		\\
        \midrule

        \multicolumn{4}{c}{\textbf{U(OOD)I(ID)}}
        \\
        \midrule
        \midrule

        Vanilla (ReACT)

        & 3.29
        & 3.28
        & 1.57
        
		\\
        % \midrule
  
		\quad FT (ReAct)
  
        & \textbf{4.09} ($\uparrow$ 24.3\%)
        & \textbf{3.75} ($\uparrow$ 14.3\%)
        & \textbf{2.79} ($\uparrow$ 77.7\%) 

        \\
        \midrule
        % \midrule

        Vanilla (FC)
        % Qwen2.5-7B-Instruct (FC)
        & 3.33
        & 2.65
        & 1.07

        \\
        % \midrule

        \quad FT (FC)
 
        & 3.81 ($\uparrow$ 14.4\%) 
        & 3.49 ($\uparrow$ 31.7\%) 
        & 2.37 ($\uparrow$ 121.5\%)
        
        \\

        % \midrule

        \quad FT (FC wo r)

        & 3.48 ($\uparrow$ 4.5\%)
        & 3.24 ($\uparrow$ 22.3\%)
        & 2.19 ($\uparrow$ 104.6\%)
       
		\\
        \midrule

        \multicolumn{4}{c}{\textbf{U(OOD)I(OOD)}}
        \\
        \midrule
        \midrule

        Vanilla (ReACT)
		
        & 2.91
        & 3.08 
		& 1.43
        
		\\
        % \midrule
  
		\quad FT (ReAct)
		
        & \textbf{3.52} ($\uparrow$ 21.0\%)
        & \textbf{3.43} ($\uparrow$ 11.4\%) 
		& \textbf{2.06} ($\uparrow$ 44.1\%)

        \\
        \midrule
        % \midrule

        Vanilla (FC)
		
        & 3.03
        & 2.64 
		& 1.07

        \\
        % \midrule

        \quad FT (FC)
		
        & 3.29 ($\uparrow$ 8.5\%)
        & 3.14 ($\uparrow$ 18.9\%) 
		& 1.73 ($\uparrow$ 61.7\%)
        
        \\

        % \midrule

        \quad FT (FC wo r)
		
        & 2.73 ($\downarrow$ 9.9\%)
        & 2.77 ($\uparrow$ 4.9\%) 
		& 1.64 ($\uparrow$ 53.3\%)
       
		\\
        \midrule

		% \bottomrule
	\end{tabular}
}
}
	\caption{The results of the Qwen2.5-7B-Instruct model fine-tuned on a subset of instructions. \textbf{U} represents the User, \textbf{I} represents the Instruction. \textbf{ID} indicates the users or instructions are seen in the training data, \textbf{OOD} means not seen in the fine-tuning. ``wo r'' means outputting without reasoning process and directly invoking tools.
	}
	\label{table:fine-tuning-result}
\end{table}

\subsection{Effectiveness of key-point-based LLM Evaluation}
\begin{figure*}[htbp]
    \centering
    \begin{minipage}[t]{0.48\textwidth}
        \centering
        \includegraphics[width=\linewidth]{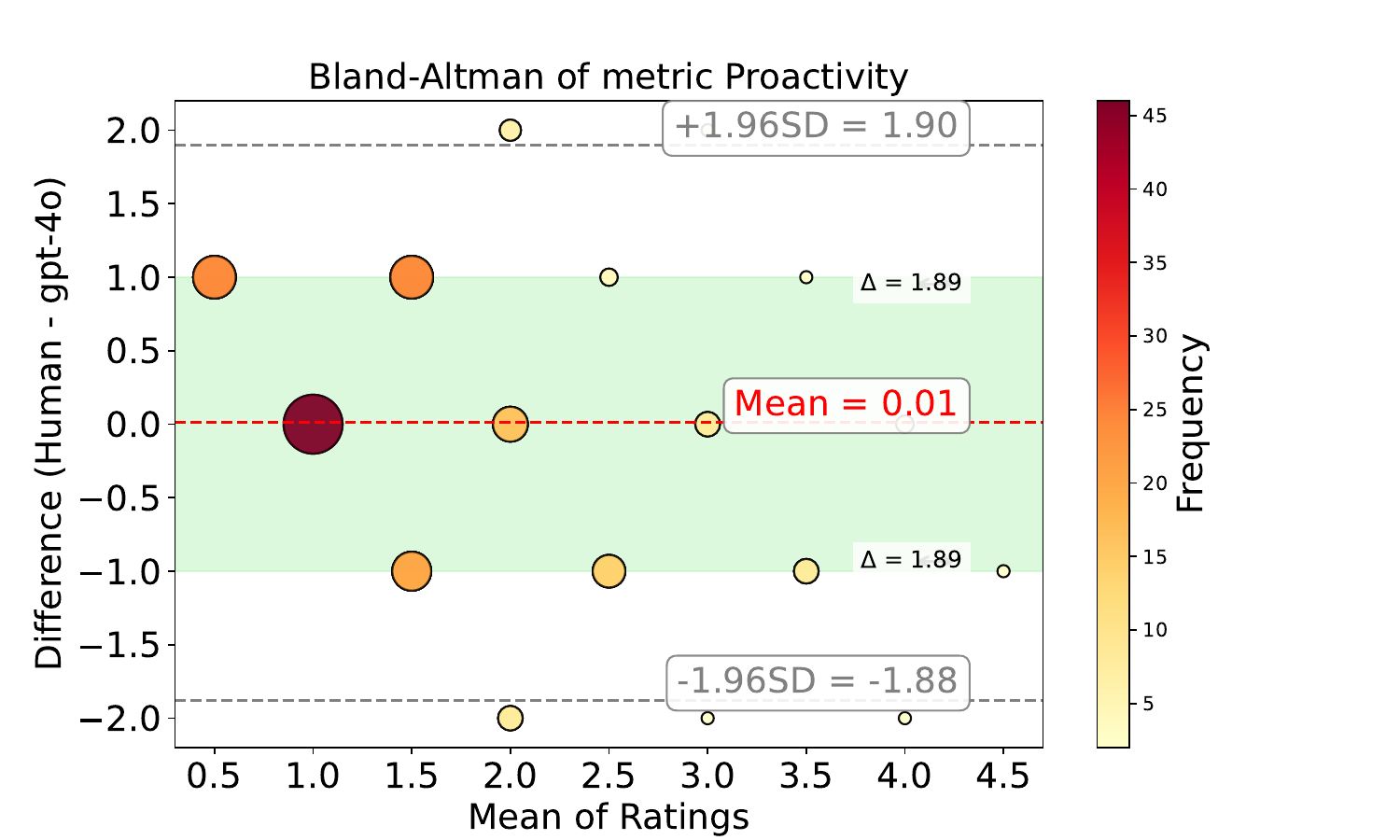}
        \caption{The Bland-Altman analysis result of \textbf{Proactivity} with given key points.}
        \label{fig:w_keypointres_proactivity}
    \end{minipage}
    \hfill 
    \begin{minipage}[t]{0.48\textwidth}
        \centering
        \includegraphics[width=\linewidth]{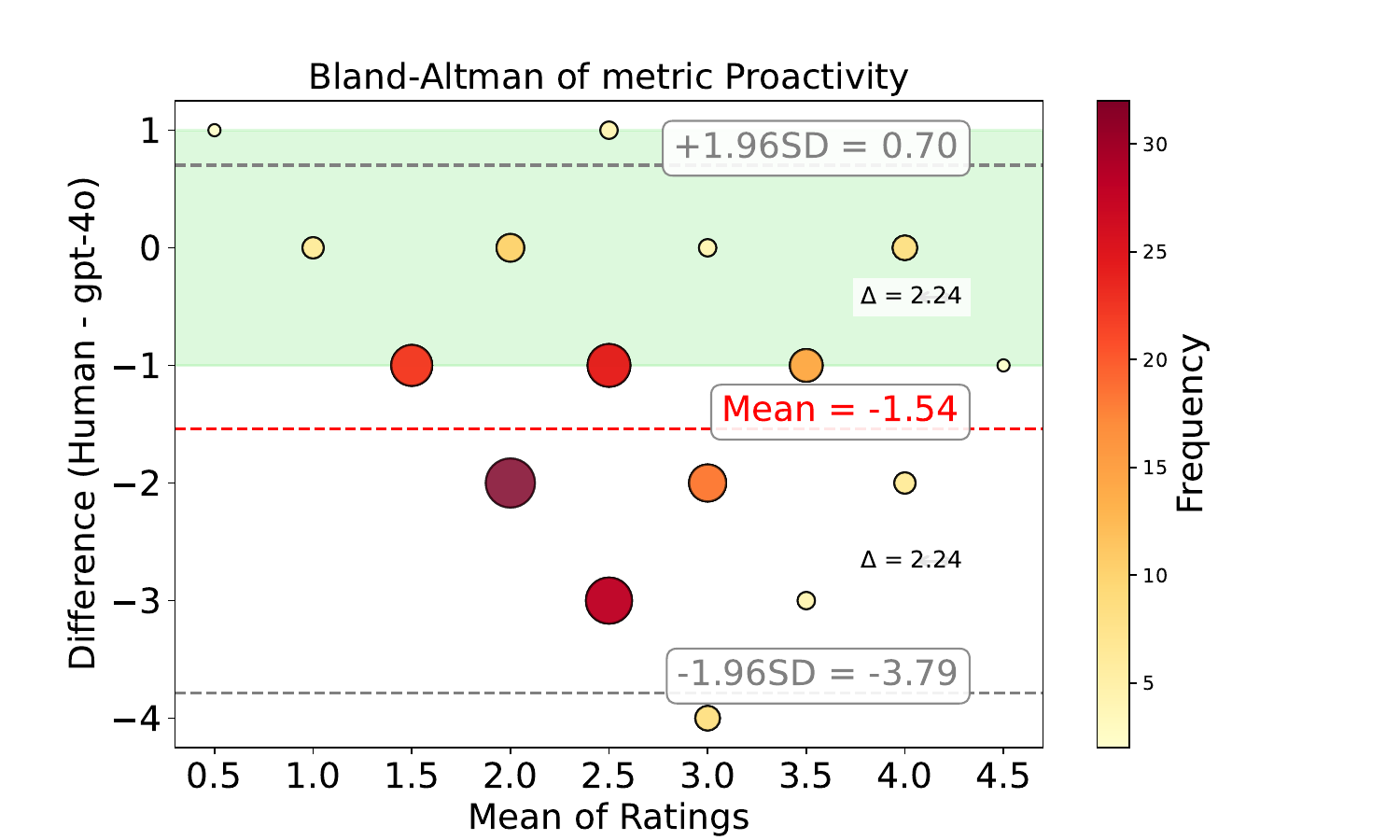}
        \caption{The Bland-Altman analysis result of \textbf{Proactivity} without given key points.}
        \label{fig:wo_keypointres_proactivity}
    \end{minipage}
\end{figure*}

To demonstrate the effectiveness of the key-point-based LLM evaluation method, we collect 16 data points per method in both Tool-Given and Tool-Retrieval settings for model GPT-4o. This result in a total of 96 data points. We then analyze the consistency with human evaluation to validate the method’s effectiveness.
% z

To quantify consistency, we use the Bland-Altman method, which plots the average of two measurements on the x-axis and their difference on the y-axis. Three reference lines are included: the mean difference (red dashed line) and the 95\% limits of agreement (LoA), calculated as mean ± 1.96 × standard deviation. If most of data points fall within the LoA, the methods show good consistency. The point size indicates sample count, and the green area highlights samples with a score difference within 1 point.

The experimental results of metric Proactivity in Figure~\ref{fig:w_keypointres_proactivity} and Figure~\ref{fig:wo_keypointres_proactivity} showed: (1) System Bias Control: Under the key-point constraint, the mean difference is 0.01 (95\% CI: -1.88 $\sim$ 1.90), close to the theoretical ideal value of 0, indicating statistical consistency between LLM and human evaluation in the total score dimension. (2) LoA Interval Length: Compared to the free evaluation mode, the key-point constraint reduces the width of the LoA interval (1.89 vs. 2.24). (3) Error Distribution Characteristics: 89.6\% of the samples show a score difference of $\leq$ 1 point (green area), representing an improvement of over 39.6 percentage points compared to the control group without key points.
These findings confirm the effectiveness of the key-point-based LLM evaluation strategy.
Please refer to Appendix~\ref{sec:appendix_effectiveness_of_key-point-based_LLM_evaluation} for more details.

\section{Related Works}

\subsection{Personalization of LLM}
The personalization of LLMs aims to align with individual preferences \cite{zollo2025personalllm, tseng2024talespersonallmssurvey}. Current personalization tasks are typically focused on specific text generation \cite{he2025cosenhancingpersonalizationmitigating, zhao2025do, lee2024aligningthousandspreferencesmessage}, recommendation tasks \cite{yang2023palrpersonalizationawarellms}, role-playing \cite{wang-etal-2024-rolellm, shao-etal-2023-character}, news headline generation \cite{salemi-etal-2024-lamp}, conversational tasks, as well as applications in education \cite{Pratama2023REVOLUTIONIZINGEH} and healthcare \cite{10.1145/3437963.3441657, abbasian2024conversationalhealthagentspersonalized}, and the evaluation process typically involves summarizing user preferences from the context to generate responses \cite{zhang2024guidedprofilegenerationimproves}. However, the Personal LLM Agents goes further by defining an ideal personal assistant that should possess interactive capabilities \cite{singh-etal-2024-personal}. We propose ETAPP to address the problem that the evaluation of Personal LLM Agents lacks criteria for personalized tool usage in diverse scenarios.

\subsection{Tool-augmented LLM}
Tool-augmented LLMs aim to enable them to interact with the external world through tools \cite{10.1145/3704435, qu2025tool}. Numerous benchmarks and methods have been proposed to evaluate and improve the accuracy of tool utilization and task completion \cite{trivedi-etal-2024-appworld,li-etal-2023-api, hao2024citienhancingtoolutilizing, tang2023toolalpacageneralizedtoollearning, NEURIPS2023_9cb2a749}. However, these datasets are designed for the general human population rather than for personalized tool invocation tailored to individual user preferences. Our work integrates the tool invocation process with personalization and proposes evaluation metrics and methods for this purpose.

\section{Conclusion}
The paper evaluates the ability and level of large language models to provide personalized services in interactive environments. We integrate the process of tool invocation with personalized services and propose two evaluation metrics: Personalization and Proactivity. To address the unreliability of using LLMs as evaluators in this task, we introduce a key-point-based evaluation method for LLMs and validate its reliability. Finally, we analyze the performance of different models on this task and explore the challenges and potential directions for improving their personalized tool utilization capabilities.

\section*{Limitations}
We evaluate the personalization of tool utilization in LLMs from the perspectives of personalization and proactivity, and propose a key-point-based LLM evaluation method. However, there are still challenges in this area. For example, our work does not consider multimodal tasks. And there are details that can be further refined. For example, during preference modeling, we could further model the selection preferences for similar tools (e.g., when two apps A and B provide the same functionality, users tend to prefer the API from App A to complete the task). In the future, we plan to extend our work to the multimodal domain.

\section*{Ethics Statement}
Our work does not introduce ethical concerns. This paper utilized AI assistance for language polishing of the manuscript, including vocabulary correction and spell checking.

\bibliography{custom}

\begin{thebibliography}{39}
\providecommand{\natexlab}[1]{#1}

\bibitem[{Abbasian et~al.(2024)Abbasian, Azimi, Rahmani, and Jain}]{abbasian2024conversationalhealthagentspersonalized}
Mahyar Abbasian, Iman Azimi, Amir~M. Rahmani, and Ramesh Jain. 2024.
\newblock \href {https://arxiv.org/abs/2310.02374} {Conversational health agents: A personalized llm-powered agent framework}.
\newblock \emph{Preprint}, arXiv:2310.02374.

\bibitem[{Cai et~al.(2024)Cai, Li, Wang, Zhu, Shen, Li, and Chua}]{cai2024largelanguagemodelsempowered}
Hongru Cai, Yongqi Li, Wenjie Wang, Fengbin Zhu, Xiaoyu Shen, Wenjie Li, and Tat-Seng Chua. 2024.
\newblock \href {https://arxiv.org/abs/2410.17236} {Large language models empowered personalized web agents}.
\newblock \emph{Preprint}, arXiv:2410.17236.

\bibitem[{Chen et~al.(2024)Chen, Wang, Xu, Yuan, Zhang, Shi, Xie, Li, Yang, Zhu, Chen, Li, Chen, Hu, Wu, Ren, Fu, and Xiao}]{chen2024personapersonalizationsurveyroleplaying}
Jiangjie Chen, Xintao Wang, Rui Xu, Siyu Yuan, Yikai Zhang, Wei Shi, Jian Xie, Shuang Li, Ruihan Yang, Tinghui Zhu, Aili Chen, Nianqi Li, Lida Chen, Caiyu Hu, Siye Wu, Scott Ren, Ziquan Fu, and Yanghua Xiao. 2024.
\newblock \href {https://arxiv.org/abs/2404.18231} {From persona to personalization: A survey on role-playing language agents}.
\newblock \emph{Preprint}, arXiv:2404.18231.

\bibitem[{DeepSeek-AI et~al.(2025)DeepSeek-AI, Guo, Yang, Zhang, Song, Zhang, Xu, Zhu, Ma, Wang, Bi, Zhang, Yu, Wu, Wu, Gou, Shao, Li, Gao, Liu, Xue, Wang, Wu, Feng, Lu, Zhao, Deng, Zhang, Ruan, Dai, Chen, Ji, Li, Lin, Dai, Luo, Hao, Chen, Li, Zhang, Bao, Xu, Wang, Ding, Xin, Gao, Qu, Li, Guo, Li, Wang, Chen, Yuan, Qiu, Li, Cai, Ni, Liang, Chen, Dong, Hu, Gao, Guan, Huang, Yu, Wang, Zhang, Zhao, Wang, Zhang, Xu, Xia, Zhang, Zhang, Tang, Li, Wang, Li, Tian, Huang, Zhang, Wang, Chen, Du, Ge, Zhang, Pan, Wang, Chen, Jin, Chen, Lu, Zhou, Chen, Ye, Wang, Yu, Zhou, Pan, Li, Zhou, Wu, Ye, Yun, Pei, Sun, Wang, Zeng, Zhao, Liu, Liang, Gao, Yu, Zhang, Xiao, An, Liu, Wang, Chen, Nie, Cheng, Liu, Xie, Liu, Yang, Li, Su, Lin, Li, Jin, Shen, Chen, Sun, Wang, Song, Zhou, Wang, Shan, Li, Wang, Wei, Zhang, Xu, Li, Zhao, Sun, Wang, Yu, Zhang, Shi, Xiong, He, Piao, Wang, Tan, Ma, Liu, Guo, Ou, Wang, Gong, Zou, He, Xiong, Luo, You, Liu, Zhou, Zhu, Xu, Huang, Li, Zheng, Zhu, Ma, Tang, Zha, Yan, Ren, Ren, Sha, Fu, Xu, Xie, Zhang,
  Hao, Ma, Yan, Wu, Gu, Zhu, Liu, Li, Xie, Song, Pan, Huang, Xu, Zhang, and Zhang}]{deepseekai2025deepseekr1incentivizingreasoningcapability}
DeepSeek-AI, Daya Guo, Dejian Yang, Haowei Zhang, Junxiao Song, Ruoyu Zhang, Runxin Xu, Qihao Zhu, Shirong Ma, Peiyi Wang, Xiao Bi, Xiaokang Zhang, Xingkai Yu, Yu~Wu, Z.~F. Wu, Zhibin Gou, Zhihong Shao, Zhuoshu Li, Ziyi Gao, Aixin Liu, Bing Xue, Bingxuan Wang, Bochao Wu, Bei Feng, Chengda Lu, Chenggang Zhao, Chengqi Deng, Chenyu Zhang, Chong Ruan, Damai Dai, Deli Chen, Dongjie Ji, Erhang Li, Fangyun Lin, Fucong Dai, Fuli Luo, Guangbo Hao, Guanting Chen, Guowei Li, H.~Zhang, Han Bao, Hanwei Xu, Haocheng Wang, Honghui Ding, Huajian Xin, Huazuo Gao, Hui Qu, Hui Li, Jianzhong Guo, Jiashi Li, Jiawei Wang, Jingchang Chen, Jingyang Yuan, Junjie Qiu, Junlong Li, J.~L. Cai, Jiaqi Ni, Jian Liang, Jin Chen, Kai Dong, Kai Hu, Kaige Gao, Kang Guan, Kexin Huang, Kuai Yu, Lean Wang, Lecong Zhang, Liang Zhao, Litong Wang, Liyue Zhang, Lei Xu, Leyi Xia, Mingchuan Zhang, Minghua Zhang, Minghui Tang, Meng Li, Miaojun Wang, Mingming Li, Ning Tian, Panpan Huang, Peng Zhang, Qiancheng Wang, Qinyu Chen, Qiushi Du, Ruiqi Ge, Ruisong
  Zhang, Ruizhe Pan, Runji Wang, R.~J. Chen, R.~L. Jin, Ruyi Chen, Shanghao Lu, Shangyan Zhou, Shanhuang Chen, Shengfeng Ye, Shiyu Wang, Shuiping Yu, Shunfeng Zhou, Shuting Pan, S.~S. Li, Shuang Zhou, Shaoqing Wu, Shengfeng Ye, Tao Yun, Tian Pei, Tianyu Sun, T.~Wang, Wangding Zeng, Wanjia Zhao, Wen Liu, Wenfeng Liang, Wenjun Gao, Wenqin Yu, Wentao Zhang, W.~L. Xiao, Wei An, Xiaodong Liu, Xiaohan Wang, Xiaokang Chen, Xiaotao Nie, Xin Cheng, Xin Liu, Xin Xie, Xingchao Liu, Xinyu Yang, Xinyuan Li, Xuecheng Su, Xuheng Lin, X.~Q. Li, Xiangyue Jin, Xiaojin Shen, Xiaosha Chen, Xiaowen Sun, Xiaoxiang Wang, Xinnan Song, Xinyi Zhou, Xianzu Wang, Xinxia Shan, Y.~K. Li, Y.~Q. Wang, Y.~X. Wei, Yang Zhang, Yanhong Xu, Yao Li, Yao Zhao, Yaofeng Sun, Yaohui Wang, Yi~Yu, Yichao Zhang, Yifan Shi, Yiliang Xiong, Ying He, Yishi Piao, Yisong Wang, Yixuan Tan, Yiyang Ma, Yiyuan Liu, Yongqiang Guo, Yuan Ou, Yuduan Wang, Yue Gong, Yuheng Zou, Yujia He, Yunfan Xiong, Yuxiang Luo, Yuxiang You, Yuxuan Liu, Yuyang Zhou, Y.~X. Zhu,
  Yanhong Xu, Yanping Huang, Yaohui Li, Yi~Zheng, Yuchen Zhu, Yunxian Ma, Ying Tang, Yukun Zha, Yuting Yan, Z.~Z. Ren, Zehui Ren, Zhangli Sha, Zhe Fu, Zhean Xu, Zhenda Xie, Zhengyan Zhang, Zhewen Hao, Zhicheng Ma, Zhigang Yan, Zhiyu Wu, Zihui Gu, Zijia Zhu, Zijun Liu, Zilin Li, Ziwei Xie, Ziyang Song, Zizheng Pan, Zhen Huang, Zhipeng Xu, Zhongyu Zhang, and Zhen Zhang. 2025.
\newblock \href {https://arxiv.org/abs/2501.12948} {Deepseek-r1: Incentivizing reasoning capability in llms via reinforcement learning}.
\newblock \emph{Preprint}, arXiv:2501.12948.

\bibitem[{DeepSeek-AI et~al.(2024)DeepSeek-AI, Liu, Feng, Xue, Wang, Wu, Lu, Zhao, Deng, Zhang, Ruan, Dai, Guo, Yang, Chen, Ji, Li, Lin, Dai, Luo, Hao, Chen, Li, Zhang, Bao, Xu, Wang, Zhang, Ding, Xin, Gao, Li, Qu, Cai, Liang, Guo, Ni, Li, Wang, Chen, Chen, Yuan, Qiu, Li, Song, Dong, Hu, Gao, Guan, Huang, Yu, Wang, Zhang, Xu, Xia, Zhao, Wang, Zhang, Li, Wang, Zhang, Zhang, Tang, Li, Tian, Huang, Wang, Zhang, Wang, Zhu, Chen, Du, Chen, Jin, Ge, Zhang, Pan, Wang, Xu, Zhang, Chen, Li, Lu, Zhou, Chen, Wu, Ye, Ye, Ma, Wang, Zhou, Yu, Zhou, Pan, Wang, Yun, Pei, Sun, Xiao, Zeng, Zhao, An, Liu, Liang, Gao, Yu, Zhang, Li, Jin, Wang, Bi, Liu, Wang, Shen, Chen, Zhang, Chen, Nie, Sun, Wang, Cheng, Liu, Xie, Liu, Yu, Song, Shan, Zhou, Yang, Li, Su, Lin, Li, Wang, Wei, Zhu, Zhang, Xu, Xu, Huang, Li, Zhao, Sun, Li, Wang, Yu, Zheng, Zhang, Shi, Xiong, He, Tang, Piao, Wang, Tan, Ma, Liu, Guo, Wu, Ou, Zhu, Wang, Gong, Zou, He, Zha, Xiong, Ma, Yan, Luo, You, Liu, Zhou, Wu, Ren, Ren, Sha, Fu, Xu, Huang, Zhang, Xie, Zhang, Hao,
  Gou, Ma, Yan, Shao, Xu, Wu, Zhang, Li, Gu, Zhu, Liu, Li, Xie, Song, Gao, and Pan}]{deepseekai2024deepseekv3technicalreport}
DeepSeek-AI, Aixin Liu, Bei Feng, Bing Xue, Bingxuan Wang, Bochao Wu, Chengda Lu, Chenggang Zhao, Chengqi Deng, Chenyu Zhang, Chong Ruan, Damai Dai, Daya Guo, Dejian Yang, Deli Chen, Dongjie Ji, Erhang Li, Fangyun Lin, Fucong Dai, Fuli Luo, Guangbo Hao, Guanting Chen, Guowei Li, H.~Zhang, Han Bao, Hanwei Xu, Haocheng Wang, Haowei Zhang, Honghui Ding, Huajian Xin, Huazuo Gao, Hui Li, Hui Qu, J.~L. Cai, Jian Liang, Jianzhong Guo, Jiaqi Ni, Jiashi Li, Jiawei Wang, Jin Chen, Jingchang Chen, Jingyang Yuan, Junjie Qiu, Junlong Li, Junxiao Song, Kai Dong, Kai Hu, Kaige Gao, Kang Guan, Kexin Huang, Kuai Yu, Lean Wang, Lecong Zhang, Lei Xu, Leyi Xia, Liang Zhao, Litong Wang, Liyue Zhang, Meng Li, Miaojun Wang, Mingchuan Zhang, Minghua Zhang, Minghui Tang, Mingming Li, Ning Tian, Panpan Huang, Peiyi Wang, Peng Zhang, Qiancheng Wang, Qihao Zhu, Qinyu Chen, Qiushi Du, R.~J. Chen, R.~L. Jin, Ruiqi Ge, Ruisong Zhang, Ruizhe Pan, Runji Wang, Runxin Xu, Ruoyu Zhang, Ruyi Chen, S.~S. Li, Shanghao Lu, Shangyan Zhou, Shanhuang
  Chen, Shaoqing Wu, Shengfeng Ye, Shengfeng Ye, Shirong Ma, Shiyu Wang, Shuang Zhou, Shuiping Yu, Shunfeng Zhou, Shuting Pan, T.~Wang, Tao Yun, Tian Pei, Tianyu Sun, W.~L. Xiao, Wangding Zeng, Wanjia Zhao, Wei An, Wen Liu, Wenfeng Liang, Wenjun Gao, Wenqin Yu, Wentao Zhang, X.~Q. Li, Xiangyue Jin, Xianzu Wang, Xiao Bi, Xiaodong Liu, Xiaohan Wang, Xiaojin Shen, Xiaokang Chen, Xiaokang Zhang, Xiaosha Chen, Xiaotao Nie, Xiaowen Sun, Xiaoxiang Wang, Xin Cheng, Xin Liu, Xin Xie, Xingchao Liu, Xingkai Yu, Xinnan Song, Xinxia Shan, Xinyi Zhou, Xinyu Yang, Xinyuan Li, Xuecheng Su, Xuheng Lin, Y.~K. Li, Y.~Q. Wang, Y.~X. Wei, Y.~X. Zhu, Yang Zhang, Yanhong Xu, Yanhong Xu, Yanping Huang, Yao Li, Yao Zhao, Yaofeng Sun, Yaohui Li, Yaohui Wang, Yi~Yu, Yi~Zheng, Yichao Zhang, Yifan Shi, Yiliang Xiong, Ying He, Ying Tang, Yishi Piao, Yisong Wang, Yixuan Tan, Yiyang Ma, Yiyuan Liu, Yongqiang Guo, Yu~Wu, Yuan Ou, Yuchen Zhu, Yuduan Wang, Yue Gong, Yuheng Zou, Yujia He, Yukun Zha, Yunfan Xiong, Yunxian Ma, Yuting Yan, Yuxiang
  Luo, Yuxiang You, Yuxuan Liu, Yuyang Zhou, Z.~F. Wu, Z.~Z. Ren, Zehui Ren, Zhangli Sha, Zhe Fu, Zhean Xu, Zhen Huang, Zhen Zhang, Zhenda Xie, Zhengyan Zhang, Zhewen Hao, Zhibin Gou, Zhicheng Ma, Zhigang Yan, Zhihong Shao, Zhipeng Xu, Zhiyu Wu, Zhongyu Zhang, Zhuoshu Li, Zihui Gu, Zijia Zhu, Zijun Liu, Zilin Li, Ziwei Xie, Ziyang Song, Ziyi Gao, and Zizheng Pan. 2024.
\newblock \href {https://arxiv.org/abs/2412.19437} {Deepseek-v3 technical report}.
\newblock \emph{Preprint}, arXiv:2412.19437.

\bibitem[{Goldenberg et~al.(2021)Goldenberg, Kofman, Albert, Mizrachi, Horowitz, and Teinemaa}]{10.1145/3437963.3441657}
Dmitri Goldenberg, Kostia Kofman, Javier Albert, Sarai Mizrachi, Adam Horowitz, and Irene Teinemaa. 2021.
\newblock \href {https://doi.org/10.1145/3437963.3441657} {Personalization in practice: Methods and applications}.
\newblock In \emph{Proceedings of the 14th ACM International Conference on Web Search and Data Mining}, WSDM '21, page 1123–1126, New York, NY, USA. Association for Computing Machinery.

\bibitem[{Hao et~al.(2024)Hao, Cao, Jin, Liao, Chen, Liu, and Zhao}]{hao2024citienhancingtoolutilizing}
Yupu Hao, Pengfei Cao, Zhuoran Jin, Huanxuan Liao, Yubo Chen, Kang Liu, and Jun Zhao. 2024.
\newblock \href {https://arxiv.org/abs/2409.13202} {Citi: Enhancing tool utilizing ability in large language models without sacrificing general performance}.
\newblock \emph{Preprint}, arXiv:2409.13202.

\bibitem[{He et~al.(2025)He, Pandey, Schrum, and Dragan}]{he2025cosenhancingpersonalizationmitigating}
Jerry Zhi-Yang He, Sashrika Pandey, Mariah~L. Schrum, and Anca Dragan. 2025.
\newblock \href {https://arxiv.org/abs/2405.01768} {Cos: Enhancing personalization and mitigating bias with context steering}.
\newblock \emph{Preprint}, arXiv:2405.01768.

\bibitem[{Jang et~al.(2023)Jang, Kim, Lin, Wang, Hessel, Zettlemoyer, Hajishirzi, Choi, and Ammanabrolu}]{jang2023personalized}
Joel Jang, Seungone Kim, Bill~Yuchen Lin, Yizhong Wang, Jack Hessel, Luke Zettlemoyer, Hannaneh Hajishirzi, Yejin Choi, and Prithviraj Ammanabrolu. 2023.
\newblock Personalized soups: Personalized large language model alignment via post-hoc parameter merging.
\newblock \emph{arXiv preprint arXiv:2310.11564}.

\bibitem[{Lee et~al.(2024)Lee, Park, Kim, and Seo}]{lee2024aligningthousandspreferencesmessage}
Seongyun Lee, Sue~Hyun Park, Seungone Kim, and Minjoon Seo. 2024.
\newblock \href {https://arxiv.org/abs/2405.17977} {Aligning to thousands of preferences via system message generalization}.
\newblock \emph{Preprint}, arXiv:2405.17977.

\bibitem[{Li et~al.(2025)Li, Jiang, Huang, Beigi, Zhao, Tan, Bhattacharjee, Jiang, Chen, Wu, Shu, Cheng, and Liu}]{li2025generationjudgmentopportunitieschallenges}
Dawei Li, Bohan Jiang, Liangjie Huang, Alimohammad Beigi, Chengshuai Zhao, Zhen Tan, Amrita Bhattacharjee, Yuxuan Jiang, Canyu Chen, Tianhao Wu, Kai Shu, Lu~Cheng, and Huan Liu. 2025.
\newblock \href {https://arxiv.org/abs/2411.16594} {From generation to judgment: Opportunities and challenges of llm-as-a-judge}.
\newblock \emph{Preprint}, arXiv:2411.16594.

\bibitem[{Li et~al.(2024{\natexlab{a}})Li, Dong, Chen, Su, Zhou, Ai, Ye, and Liu}]{li2024llmsasjudgescomprehensivesurveyllmbased}
Haitao Li, Qian Dong, Junjie Chen, Huixue Su, Yujia Zhou, Qingyao Ai, Ziyi Ye, and Yiqun Liu. 2024{\natexlab{a}}.
\newblock \href {https://arxiv.org/abs/2412.05579} {Llms-as-judges: A comprehensive survey on llm-based evaluation methods}.
\newblock \emph{Preprint}, arXiv:2412.05579.

\bibitem[{Li et~al.(2023)Li, Zhao, Yu, Song, Li, Yu, Li, Huang, and Li}]{li-etal-2023-api}
Minghao Li, Yingxiu Zhao, Bowen Yu, Feifan Song, Hangyu Li, Haiyang Yu, Zhoujun Li, Fei Huang, and Yongbin Li. 2023.
\newblock \href {https://doi.org/10.18653/v1/2023.emnlp-main.187} {{API}-bank: A comprehensive benchmark for tool-augmented {LLM}s}.
\newblock In \emph{Proceedings of the 2023 Conference on Empirical Methods in Natural Language Processing}, pages 3102--3116, Singapore. Association for Computational Linguistics.

\bibitem[{Li et~al.(2024{\natexlab{b}})Li, Wen, Wang, Li, Yuan, Liu, Liu, Xu, Wang, Sun, Kong, Wang, Geng, Luan, Jin, Ye, Xiong, Zhang, Li, Xu, Li, Li, Liu, Zhang, and Liu}]{li2024personalllmagentsinsights}
Yuanchun Li, Hao Wen, Weijun Wang, Xiangyu Li, Yizhen Yuan, Guohong Liu, Jiacheng Liu, Wenxing Xu, Xiang Wang, Yi~Sun, Rui Kong, Yile Wang, Hanfei Geng, Jian Luan, Xuefeng Jin, Zilong Ye, Guanjing Xiong, Fan Zhang, Xiang Li, Mengwei Xu, Zhijun Li, Peng Li, Yang Liu, Ya-Qin Zhang, and Yunxin Liu. 2024{\natexlab{b}}.
\newblock \href {https://arxiv.org/abs/2401.05459} {Personal llm agents: Insights and survey about the capability, efficiency and security}.
\newblock \emph{Preprint}, arXiv:2401.05459.

\bibitem[{Mialon et~al.(2023)Mialon, Dessì, Lomeli, Nalmpantis, Pasunuru, Raileanu, Rozière, Schick, Dwivedi-Yu, Celikyilmaz, Grave, LeCun, and Scialom}]{mialon2023augmentedlanguagemodelssurvey}
Grégoire Mialon, Roberto Dessì, Maria Lomeli, Christoforos Nalmpantis, Ram Pasunuru, Roberta Raileanu, Baptiste Rozière, Timo Schick, Jane Dwivedi-Yu, Asli Celikyilmaz, Edouard Grave, Yann LeCun, and Thomas Scialom. 2023.
\newblock \href {https://arxiv.org/abs/2302.07842} {Augmented language models: a survey}.
\newblock \emph{Preprint}, arXiv:2302.07842.

\bibitem[{Pratama et~al.(2023)Pratama, Sampelolo, and Lura}]{Pratama2023REVOLUTIONIZINGEH}
Muh.~Putra Pratama, Rigel Sampelolo, and Hans Lura. 2023.
\newblock \href {https://api.semanticscholar.org/CorpusID:268216550} {Revolutionizing education: Harnessing the power of artificial intelligence for personalized learning}.
\newblock \emph{KLASIKAL : JOURNAL OF EDUCATION, LANGUAGE TEACHING AND SCIENCE}.

\bibitem[{Qin et~al.(2024{\natexlab{a}})Qin, Hu, Lin, Chen, Ding, Cui, Zeng, Zhou, Huang, Xiao, Han, Fung, Su, Wang, Qian, Tian, Zhu, Liang, Shen, Xu, Zhang, Ye, Li, Tang, Yi, Zhu, Dai, Yan, Cong, Lu, Zhao, Huang, Yan, Han, Sun, Li, Phang, Yang, Wu, Ji, Li, Liu, and Sun}]{10.1145/3704435}
Yujia Qin, Shengding Hu, Yankai Lin, Weize Chen, Ning Ding, Ganqu Cui, Zheni Zeng, Xuanhe Zhou, Yufei Huang, Chaojun Xiao, Chi Han, Yi~Ren Fung, Yusheng Su, Huadong Wang, Cheng Qian, Runchu Tian, Kunlun Zhu, Shihao Liang, Xingyu Shen, Bokai Xu, Zhen Zhang, Yining Ye, Bowen Li, Ziwei Tang, Jing Yi, Yuzhang Zhu, Zhenning Dai, Lan Yan, Xin Cong, Yaxi Lu, Weilin Zhao, Yuxiang Huang, Junxi Yan, Xu~Han, Xian Sun, Dahai Li, Jason Phang, Cheng Yang, Tongshuang Wu, Heng Ji, Guoliang Li, Zhiyuan Liu, and Maosong Sun. 2024{\natexlab{a}}.
\newblock \href {https://doi.org/10.1145/3704435} {Tool learning with foundation models}.
\newblock \emph{ACM Comput. Surv.}, 57(4).

\bibitem[{Qin et~al.(2024{\natexlab{b}})Qin, Liang, Ye, Zhu, Yan, Lu, Lin, Cong, Tang, Qian, Zhao, Hong, Tian, Xie, Zhou, Gerstein, Li, Liu, and Sun}]{DBLP:conf/iclr/QinLYZYLLCTQZHT24}
Yujia Qin, Shihao Liang, Yining Ye, Kunlun Zhu, Lan Yan, Yaxi Lu, Yankai Lin, Xin Cong, Xiangru Tang, Bill Qian, Sihan Zhao, Lauren Hong, Runchu Tian, Ruobing Xie, Jie Zhou, Mark Gerstein, Dahai Li, Zhiyuan Liu, and Maosong Sun. 2024{\natexlab{b}}.
\newblock \href {https://openreview.net/forum?id=dHng2O0Jjr} {Toolllm: Facilitating large language models to master 16000+ real-world apis}.
\newblock In \emph{ICLR}.

\bibitem[{Qu et~al.(2025)Qu, Dai, Wei, Cai, Wang, Yin, Xu, and Wen}]{qu2025tool}
Changle Qu, Sunhao Dai, Xiaochi Wei, Hengyi Cai, Shuaiqiang Wang, Dawei Yin, Jun Xu, and Ji-Rong Wen. 2025.
\newblock Tool learning with large language models: A survey.
\newblock \emph{Frontiers of Computer Science}, 19(8):198343.

\bibitem[{Salemi et~al.(2024)Salemi, Mysore, Bendersky, and Zamani}]{salemi-etal-2024-lamp}
Alireza Salemi, Sheshera Mysore, Michael Bendersky, and Hamed Zamani. 2024.
\newblock \href {https://doi.org/10.18653/v1/2024.acl-long.399} {{L}a{MP}: When large language models meet personalization}.
\newblock In \emph{Proceedings of the 62nd Annual Meeting of the Association for Computational Linguistics (Volume 1: Long Papers)}, pages 7370--7392, Bangkok, Thailand. Association for Computational Linguistics.

\bibitem[{Shao et~al.(2023)Shao, Li, Dai, and Qiu}]{shao-etal-2023-character}
Yunfan Shao, Linyang Li, Junqi Dai, and Xipeng Qiu. 2023.
\newblock \href {https://aclanthology.org/2023.emnlp-main.814/} {Character-{LLM}: A trainable agent for role-playing}.
\newblock In \emph{Proceedings of the 2023 Conference on Empirical Methods in Natural Language Processing}, pages 13153--13187, Singapore. Association for Computational Linguistics.

\bibitem[{Singh et~al.(2024)Singh, Verma, Wang, Bharadwaj, Fashandi, Ferreira, and Lee}]{singh-etal-2024-personal}
Harmanpreet Singh, Nikhil Verma, Yixiao Wang, Manasa Bharadwaj, Homa Fashandi, Kevin Ferreira, and Chul Lee. 2024.
\newblock \href {https://doi.org/10.18653/v1/2024.emnlp-industry.37} {Personal large language model agents: A case study on tailored travel planning}.
\newblock In \emph{Proceedings of the 2024 Conference on Empirical Methods in Natural Language Processing: Industry Track}, pages 486--514, Miami, Florida, US. Association for Computational Linguistics.

\bibitem[{Tang et~al.(2023)Tang, Deng, Lin, Han, Liang, Cao, and Sun}]{tang2023toolalpacageneralizedtoollearning}
Qiaoyu Tang, Ziliang Deng, Hongyu Lin, Xianpei Han, Qiao Liang, Boxi Cao, and Le~Sun. 2023.
\newblock \href {https://arxiv.org/abs/2306.05301} {Toolalpaca: Generalized tool learning for language models with 3000 simulated cases}.
\newblock \emph{Preprint}, arXiv:2306.05301.

\bibitem[{Team(2024{\natexlab{a}})}]{qwen2.5}
Qwen Team. 2024{\natexlab{a}}.
\newblock \href {https://qwenlm.github.io/blog/qwen2.5/} {Qwen2.5: A party of foundation models}.

\bibitem[{Team(2024{\natexlab{b}})}]{qwq-32b-preview}
Qwen Team. 2024{\natexlab{b}}.
\newblock \href {https://qwenlm.github.io/blog/qwq-32b-preview/} {Qwq: Reflect deeply on the boundaries of the unknown}.

\bibitem[{Trivedi et~al.(2024)Trivedi, Khot, Hartmann, Manku, Dong, Li, Gupta, Sabharwal, and Balasubramanian}]{trivedi-etal-2024-appworld}
Harsh Trivedi, Tushar Khot, Mareike Hartmann, Ruskin Manku, Vinty Dong, Edward Li, Shashank Gupta, Ashish Sabharwal, and Niranjan Balasubramanian. 2024.
\newblock \href {https://doi.org/10.18653/v1/2024.acl-long.850} {{A}pp{W}orld: A controllable world of apps and people for benchmarking interactive coding agents}.
\newblock In \emph{Proceedings of the 62nd Annual Meeting of the Association for Computational Linguistics (Volume 1: Long Papers)}, pages 16022--16076, Bangkok, Thailand. Association for Computational Linguistics.

\bibitem[{Tseng et~al.(2024)Tseng, Huang, Hsiao, Chen, Huang, Meng, and Chen}]{tseng2024talespersonallmssurvey}
Yu-Min Tseng, Yu-Chao Huang, Teng-Yun Hsiao, Wei-Lin Chen, Chao-Wei Huang, Yu~Meng, and Yun-Nung Chen. 2024.
\newblock \href {https://arxiv.org/abs/2406.01171} {Two tales of persona in llms: A survey of role-playing and personalization}.
\newblock \emph{Preprint}, arXiv:2406.01171.

\bibitem[{Wang et~al.(2024{\natexlab{a}})Wang, Peng, Que, Liu, Zhou, Wu, Guo, Gan, Ni, Yang, Zhang, Zhang, Ouyang, Xu, Huang, Fu, and Peng}]{wang-etal-2024-rolellm}
Noah Wang, Z.y. Peng, Haoran Que, Jiaheng Liu, Wangchunshu Zhou, Yuhan Wu, Hongcheng Guo, Ruitong Gan, Zehao Ni, Jian Yang, Man Zhang, Zhaoxiang Zhang, Wanli Ouyang, Ke~Xu, Wenhao Huang, Jie Fu, and Junran Peng. 2024{\natexlab{a}}.
\newblock \href {https://doi.org/10.18653/v1/2024.findings-acl.878} {{R}ole{LLM}: Benchmarking, eliciting, and enhancing role-playing abilities of large language models}.
\newblock In \emph{Findings of the Association for Computational Linguistics: ACL 2024}, pages 14743--14777, Bangkok, Thailand. Association for Computational Linguistics.

\bibitem[{Wang et~al.(2024{\natexlab{b}})Wang, Tao, Fang, Wang, Wang, Jiang, and Zhou}]{wang2024aipersonalifelongpersonalization}
Tiannan Wang, Meiling Tao, Ruoyu Fang, Huilin Wang, Shuai Wang, Yuchen~Eleanor Jiang, and Wangchunshu Zhou. 2024{\natexlab{b}}.
\newblock \href {https://arxiv.org/abs/2412.13103} {Ai persona: Towards life-long personalization of llms}.
\newblock \emph{Preprint}, arXiv:2412.13103.

\bibitem[{Xie et~al.(2024)Xie, Zhang, Chen, Zhu, Lou, Tian, Xiao, and Su}]{xie2024travelplanner}
Jian Xie, Kai Zhang, Jiangjie Chen, Tinghui Zhu, Renze Lou, Yuandong Tian, Yanghua Xiao, and Yu~Su. 2024.
\newblock Travelplanner: A benchmark for real-world planning with language agents.
\newblock In \emph{Forty-first International Conference on Machine Learning}.

\bibitem[{Yan et~al.(2024)Yan, Mao, Ji, Zhang, Patil, Stoica, and Gonzalez}]{berkeley-function-calling-leaderboard}
Fanjia Yan, Huanzhi Mao, Charlie Cheng-Jie Ji, Tianjun Zhang, Shishir~G. Patil, Ion Stoica, and Joseph~E. Gonzalez. 2024.
\newblock Berkeley function calling leaderboard.

\bibitem[{Yang et~al.(2023)Yang, Chen, Jiang, Cho, Huang, and Lu}]{yang2023palrpersonalizationawarellms}
Fan Yang, Zheng Chen, Ziyan Jiang, Eunah Cho, Xiaojiang Huang, and Yanbin Lu. 2023.
\newblock \href {https://arxiv.org/abs/2305.07622} {Palr: Personalization aware llms for recommendation}.
\newblock \emph{Preprint}, arXiv:2305.07622.

\bibitem[{Yao et~al.(2023)Yao, Zhao, Yu, Du, Shafran, Narasimhan, and Cao}]{yao2023react}
Shunyu Yao, Jeffrey Zhao, Dian Yu, Nan Du, Izhak Shafran, Karthik~R Narasimhan, and Yuan Cao. 2023.
\newblock \href {https://openreview.net/forum?id=WE_vluYUL-X} {React: Synergizing reasoning and acting in language models}.
\newblock In \emph{The Eleventh International Conference on Learning Representations}.

\bibitem[{Zhang(2024)}]{zhang2024guidedprofilegenerationimproves}
Jiarui Zhang. 2024.
\newblock \href {https://arxiv.org/abs/2409.13093} {Guided profile generation improves personalization with llms}.
\newblock \emph{Preprint}, arXiv:2409.13093.

\bibitem[{Zhang et~al.(2024)Zhang, Bo, Ma, Li, Chen, Dai, Zhu, Dong, and Wen}]{zhang2024surveymemorymechanismlarge}
Zeyu Zhang, Xiaohe Bo, Chen Ma, Rui Li, Xu~Chen, Quanyu Dai, Jieming Zhu, Zhenhua Dong, and Ji-Rong Wen. 2024.
\newblock \href {https://arxiv.org/abs/2404.13501} {A survey on the memory mechanism of large language model based agents}.
\newblock \emph{Preprint}, arXiv:2404.13501.

\bibitem[{Zhao et~al.(2025)Zhao, Hong, Liu, Hazarika, and Lin}]{zhao2025do}
Siyan Zhao, Mingyi Hong, Yang Liu, Devamanyu Hazarika, and Kaixiang Lin. 2025.
\newblock \href {https://openreview.net/forum?id=QWunLKbBGF} {Do {LLM}s recognize your preferences? evaluating personalized preference following in {LLM}s}.
\newblock In \emph{The Thirteenth International Conference on Learning Representations}.

\bibitem[{Zhao et~al.(2024)Zhao, Zhou, Li, Tang, Wang, Hou, Min, Zhang, Zhang, Dong, Du, Yang, Chen, Chen, Jiang, Ren, Li, Tang, Liu, Liu, Nie, and Wen}]{zhao2024surveylargelanguagemodels}
Wayne~Xin Zhao, Kun Zhou, Junyi Li, Tianyi Tang, Xiaolei Wang, Yupeng Hou, Yingqian Min, Beichen Zhang, Junjie Zhang, Zican Dong, Yifan Du, Chen Yang, Yushuo Chen, Zhipeng Chen, Jinhao Jiang, Ruiyang Ren, Yifan Li, Xinyu Tang, Zikang Liu, Peiyu Liu, Jian-Yun Nie, and Ji-Rong Wen. 2024.
\newblock \href {https://arxiv.org/abs/2303.18223} {A survey of large language models}.
\newblock \emph{Preprint}, arXiv:2303.18223.

\bibitem[{Zhuang et~al.(2023)Zhuang, Yu, Wang, Sun, and Zhang}]{NEURIPS2023_9cb2a749}
Yuchen Zhuang, Yue Yu, Kuan Wang, Haotian Sun, and Chao Zhang. 2023.
\newblock \href {https://proceedings.neurips.cc/paper_files/paper/2023/file/9cb2a7495900f8b602cb10159246a016-Paper-Datasets_and_Benchmarks.pdf} {Toolqa: A dataset for llm question answering with external tools}.
\newblock In \emph{Advances in Neural Information Processing Systems}, volume~36, pages 50117--50143. Curran Associates, Inc.

\bibitem[{Zollo et~al.(2025)Zollo, Siah, Ye, Li, and Namkoong}]{zollo2025personalllm}
Thomas~P Zollo, Andrew Wei~Tung Siah, Naimeng Ye, Ang Li, and Hongseok Namkoong. 2025.
\newblock \href {https://openreview.net/forum?id=2R7498e2Tx} {Personal{LLM}: Tailoring {LLM}s to individual preferences}.
\newblock In \emph{The Thirteenth International Conference on Learning Representations}.

\end{thebibliography}
\newpage
\appendix

\section{Dataset Construction}
\label{sec:dataset_construction}

\subsection{Tools Construction}
The tools we constructed are in Table~\ref{table:api_name}. Real-world API data are used to generate outputs for most of the tools. For example, the weather data are collected from WeatherAPI.com \footnote{https://rapidapi.com/weatherapi/api/weatherapi-com}, the music data are collected from 
Genius - Song Lyrics \footnote{https://rapidapi.com/Glavier/api/genius-song-lyrics1}  and the shopping products are collected from Real-Time Amazon Data \footnote{https://rapidapi.com/letscrape-6bRBa3QguO5/api/real-time-amazon-data} in RapidAPI. And the Navigation data are modified from TravelPlanner \cite{xie2024travelplanner} which is also collected from Real-World. We change the records' date and add dozens of data points generated by gpt-4o. And the News data are entirely generated by gpt-4o.   

\subsection{The User Preference}
We construct 16 users in our benchmark, each user has a User Profile and Tool-utilizing Preference.

The User Profile example is in Example~\ref{Example:User_Profile}. The example of Tool-utilizing Preference is in Example~\ref{Example:Tool_utilizing_Preference}. One example of the interaction history of API \textit{get\_user\_recent\_workout\_records} of user \textit{James Harrington} is in Example~\ref{Example:Interaction_History}.

\subsection{The Interaction History Construction Process}
\label{sec:appendix_instruction_history}
To ensure the consistency of the interaction history, we first build a 9-day arrangement for each user and construct an interaction history based on it, covering schedule arrangements, alarms, health status (updated hourly), exercise records, email records, and music collections. The specific construction steps for the interaction history are as follows: (1)	Generate the user’s 9-day schedule based on the user profile. (2)	Extend the schedule in detail. (3)	Generate the corresponding interaction history based on the user profile and schedule. (4)	Perform manual verification to ensure consistency and accuracy.

\subsection{Key Points Annotation}
For manually annotated key points, they are annotated and verified by two graduate students to ensure their comprehensiveness and reliability as much as possible.

\section{Evaluation details}
\label{sec:evaluation_details}

\subsection{Tool-invoking Method}
The tool-invoking prompt in FC, ReAct and E-ReAct format in Tool-Given and Tool-Retrieval setting is shown in Prompt~\ref{Prompt:ReAct_Tool_Given}, Prompt~\ref{Prompt:FC_Tool_Given}, Prompt~\ref{Prompt:E-ReAct_Tool_Given}, Prompt~\ref{Prompt:ReAct_Tool_Retrieval}, Prompt~\ref{Prompt:FC_Tool_Retrieval}, Prompt~\ref{Prompt:E-ReAct_Tool_Retrieval}. The item in ``\{'' and ``\}'' will be replaced by specific content in the inference process.

\subsection{Implementation Details}
We use the following model with the function calling ability as baselines in two experimental settings: gpt-4o-2024-11-20, DeepSeek-V3 \cite{deepseekai2024deepseekv3technicalreport}, Qwen2.5-72B-Instruct \cite{qwen2.5}, Llama-3.1-70B-Instruct \footnote{https://huggingface.co/meta-llama/Llama-3.1-70B-Instruct}, and watt-tool-70B \footnote{https://huggingface.co/watt-ai/watt-tool-70B} (the best-performing model in BFCL). 
Additionally, we evaluate the reasoning model o1-preview-2024-09-12, o1-mini-2024-09-12, DeepSeek-R1 \cite{deepseekai2025deepseekr1incentivizingreasoningcapability}, DeepSeek-R1-Distill-Qwen-32B \cite{deepseekai2025deepseekr1incentivizingreasoningcapability}, and QwQ-32B-Preview \cite{qwq-32b-preview} evaluated on a subset of the 100 testing data due to the budgets factors using ReAct method, as they do not support FC.

For the reasoning model, the reasoning content is not included in the tool-invoking process in our experiments.

For the fine-tuning experiments, we fine-tune the model with LoRA. The rank r is set to 8 and the lora alpha is 16. While the epoch is set to 3 and the learning rate is 1e-4. The batch size is set to 32. The data is in interaction format and we divide each interaction turn with APIs as training data, and we totally divide 200 instructions we annotated into 841 training data. The experiments are conducted on NVIDIA A100.

\subsection{Prompt of Evaluation}
The Prompt of key-point-based LLM Evaluation is in Prompt~\ref{Prompt:Key_Point_Evaluation}. The item in ``\{'' and ``\}'' will be replaced by specific content in the evaluation process. One evaluation output conducted by GPT-4o is shown in Example~\ref{Example:Evaluation_Result_Example}.

\subsection{Example of Error}
\label{sec:appendix_example_error}
Some error examples are shown in Example~\ref{Example:Example_of_Error}.

\section{Effectiveness of key-point-based LLM evaluation}

\label{sec:appendix_effectiveness_of_key-point-based_LLM_evaluation}
\begin{figure*}[htbp]
    \centering
    \begin{minipage}[t]{0.48\textwidth}
        \centering
        \includegraphics[width=\linewidth]{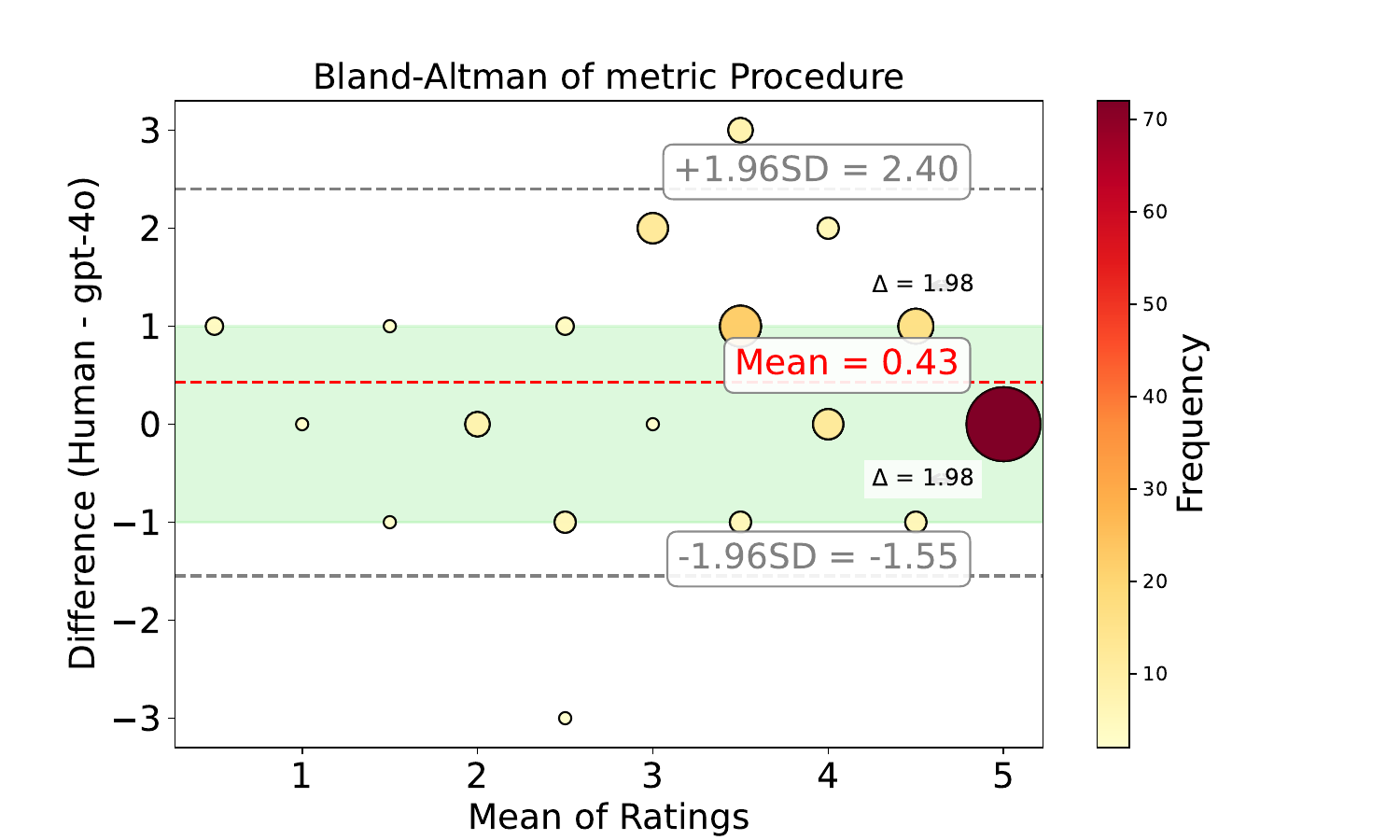}
        \caption{The Bland-Altman analysis result of \textbf{Procedure} with given key points.}
        \label{fig:w_keypointres_procedure}
    \end{minipage}
    \hfill % 添加一些水平间距
    \begin{minipage}[t]{0.48\textwidth}
        \centering
        \includegraphics[width=\linewidth]{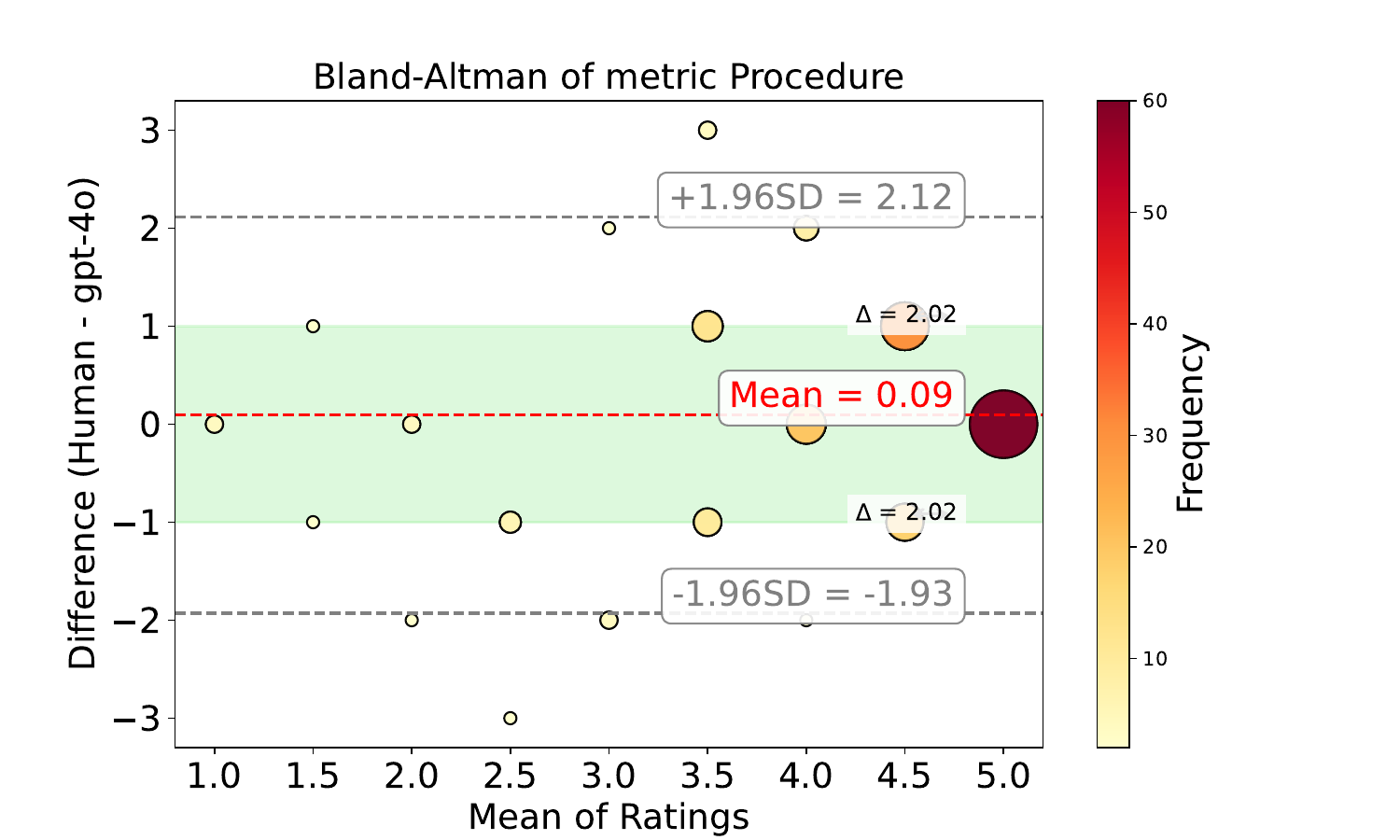}
        \caption{The Bland-Altman analysis result of \textbf{Procedure} without given key points.}
        \label{fig:wo_keypoint_res_procedure}
    \end{minipage}
\end{figure*}

\begin{figure*}[htbp]
    \centering
    \begin{minipage}[t]{0.48\textwidth}
        \centering
        \includegraphics[width=\linewidth]{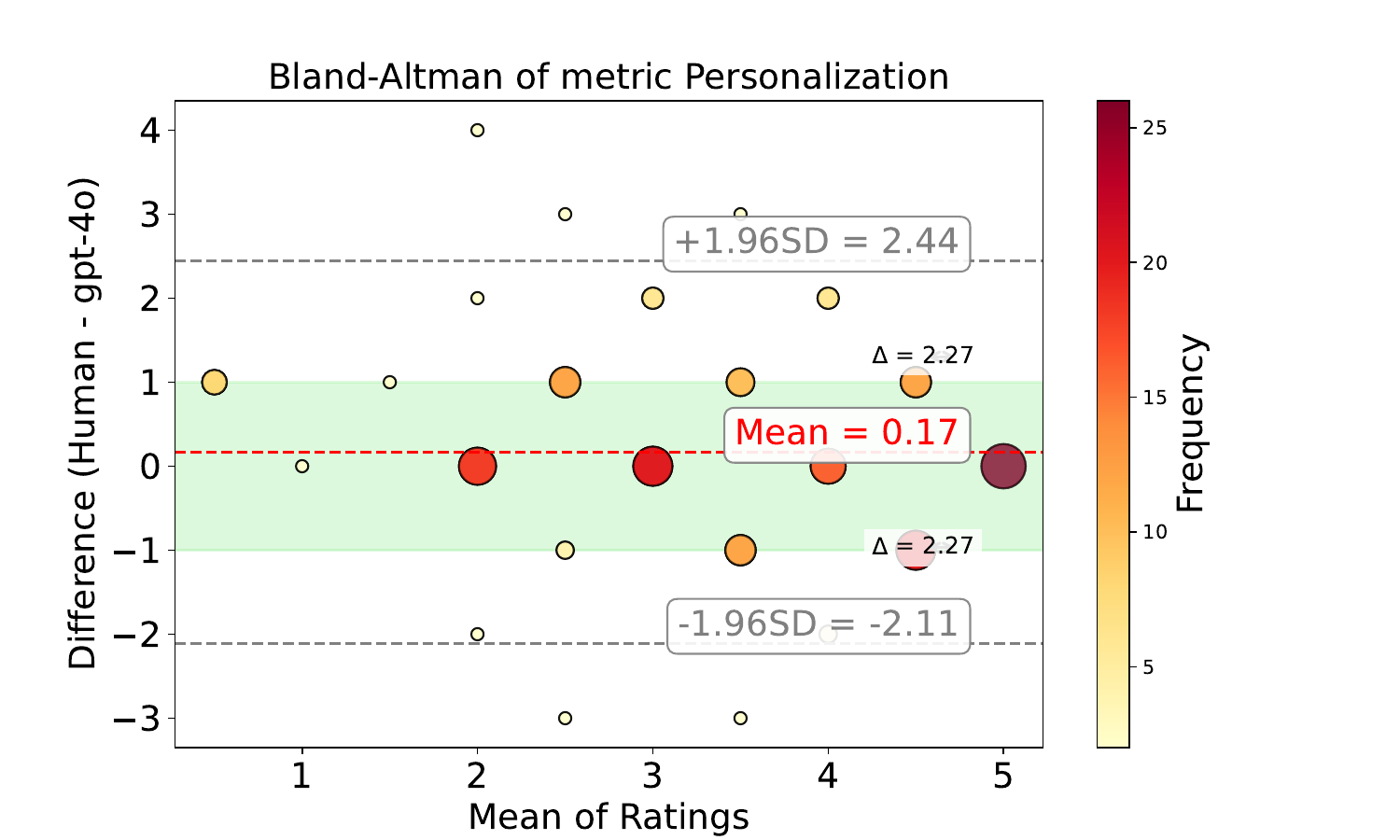}
        \caption{The Bland-Altman analysis result of \textbf{Personalization} with given key points.}
        \label{fig:w_keypoint_res_personalization}
    \end{minipage}
    \hfill % 添加一些水平间距
    \begin{minipage}[t]{0.48\textwidth}
        \centering
        \includegraphics[width=\linewidth]{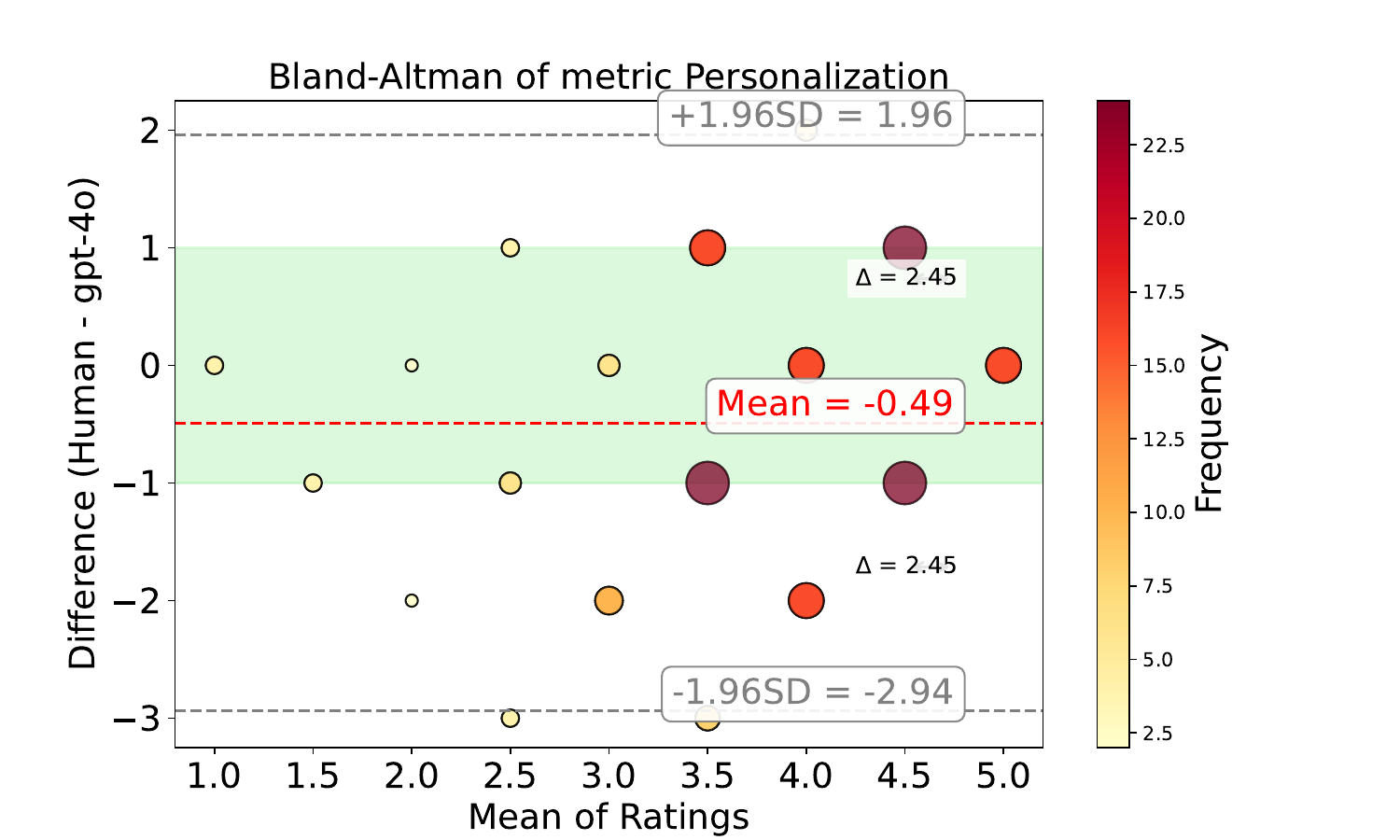}
        \caption{The Bland-Altman analysis result of \textbf{Personalization} without given key points.}
        \label{fig:wo_keypoint_res_personalization}
    \end{minipage}
\end{figure*}

The Bland-Altman analysis results of metrics Procedure and Personalization are in Figure~\ref{fig:w_keypointres_procedure}, Figure~\ref{fig:wo_keypoint_res_procedure}, Figure~\ref{fig:w_keypoint_res_personalization}, Figure~\ref{fig:wo_keypoint_res_personalization}. We observe that key points play a crucial role in improving the alignment of the evaluation process for the metrics of Personalization and Proactivity. Although the evaluation model is provided with scoring criteria, it tends to assign higher scores even without the key points. We think this may be because the model lacks sufficient knowledge and understanding of how an assistant can provide personalized and proactive services. Once the key points are input, the evaluation model can analyze whether each point is satisfied, which enhances the final analysis and the accuracy of the score. However, we also notice that the mean difference in the Procedure metric performs worse when key points are included. This may be due to the simultaneous evaluation of all three metrics, where Procedure is influenced by the other two metrics, leading to a lower score overall.

\begin{table*}
{
    \centering
    \resizebox{1\textwidth}{!}{
    \begin{tabular}{c|c}
         % &  \\
         \toprule
         Category & API Name \\
         \midrule
         HealthMonitoringApp & get\_current\_health\_and\_mood\_status \\
HealthMonitoringApp & get\_user\_recent\_workout\_records \\
HealthMonitoringApp & get\_recent\_health\_and\_mood\_summary \\
Calendar & add\_event\_in\_calendar \\
Calendar & view\_today\_events\_in\_calendar \\
Calendar & view\_events\_in\_calendar\_by\_providing\_time\_range \\
Calendar & delete\_event\_in\_calendar \\
Calendar & add\_alarm \\
Calendar & view\_today\_alarms \\
WeatherData & get\_today\_weather \\
WeatherData & get\_future\_weather \\
ECommerce & add\_product\_to\_cart \\
ECommerce & search\_products\_in\_shopping\_manager \\
ECommerce & view\_cart\_in\_shopping\_manager \\
Email & send\_email \\
Email & get\_today\_emails\_until\_now \\
Email & search\_email\_by\_sender\_and\_receiver \\
Email & search\_email\_by\_content \\
Browser & search\_news\_by\_category \\
Browser & search\_heat\_news \\
Browser & search\_from\_wikipedia \\
MusicStreamingApp & play\_music \\
MusicStreamingApp & get\_music\_list\_in\_favorites \\

Navigation\_And\_Map & find\_accommodations \\
Navigation\_And\_Map & find\_attractions \\
Navigation\_And\_Map & find\_restaurants \\
Navigation\_And\_Map & find\_flight \\
Thermostat (Smart\_Home\_Devices) & set\_temperature\_and\_humidity\_in\_home \\
Thermostat (Smart\_Home\_Devices) & get\_home\_temperature\_and\_humidity \\
Light (Smart\_Home\_Devices) & control\_light\_in\_home \\
SmartAppliance (Smart\_Home\_Devices) & control\_curtains\_in\_home \\
SmartAppliance (Smart\_Home\_Devices) & control\_bathtub\_in\_home \\
SmartAppliance (Smart\_Home\_Devices) & boil\_water\_in\_home \\
\midrule
Tool\_Searcher & search\_tools \\
Tool\_Searcher & get\_tool\_doc \\
\midrule
    \end{tabular}
    }
}
    \caption{The information of APIs and their their corresponding category.}
    \label{table:api_name}
\end{table*}

\begin{table*}[]
    \centering
    \begin{tabular}{c|c}
        \toprule
        Statistics & Number \\
        \toprule
        Tool Categories & 9 \\
        APIs Number & 35 \\
        Instruction Number & 800 \\
        User Number & 16 \\
        Key points Number of Personalization per query & 3.18 \\
        Key points Number of Proactivity per query & 3.2 \\
        Avg. Turns per query  & 3.43 \\
        \midrule
    \end{tabular}
    \caption{The statistics of ETAPP. The Avg. Turn is calculated by the output of GPT-4o in ReAct format using Tool-Given Setting.}
    \label{table:statistics}
\end{table*}

\newpage
\onecolumn
\begin{tcolorbox}[width=\textwidth, breakable]
{

\renewcommand{\lstlistingname}{Example}

\begin{lstlisting}[frame=none, basicstyle=\ttfamily\footnotesize, breaklines=true, columns=full, postbreak=\mbox{\textcolor{red}{$\hookrightarrow$}\space}, % Indicate line breaks% basicstyle=\ttfamily,    % Use monospaced font
    breakatwhitespace=true, label=Example:User_Profile, caption=User Profile Example, captionpos=b, ]
{
    "DemographicData": {
        "BasicInformation": {
            "Name": "James Harrington",
            "Age": 45,
            "Gender": "Male",
            "Location": "San Francisco",
            "Occupation": "CEO of Tech Innovations Inc.",
            "Character": "Extroverted, strategic, driven"
        },
        "SocioeconomicStatus": {
            "IncomeLevel": "High income (>$500,000 annually)",
            "EducationLevel": "MBA from Stanford University"
        },
        "CulturalBackground": {
            "LanguageProficiency": "Fluent in English and Spanish",
            "CulturalPreferences": "N/A"
        },
        "InterestActivity": "Golf, networking events, yacht sailing"
    },
    "Preferences": {
        "Entertainment": {
            "MusicGenres": "Classical, Jazz",
            "MoviesAndTVShows": "business dramas and documentaries (e.g., 'The Social Network')",
            "BooksAndAuthors": "Business leadership books by authors like Michael E. Porter, biographies of tech moguls"
        },
        "Lifestyle": {
            "DietaryHabits": "Prefers healthy and balanced meals; organic foods; follows a Mediterranean diet, Gourmet cuisine, fine dining, wine tasting, like Western cuisine and Chinese cuisine.",
            "Hobbies": "Collecting art, attending tech conferences, mentoring startups"
        },
        "Technology": {
            "FavoriteApps": "Slack, LinkedIn, Bloomberg, Evernote",
            "DevicePreferences": "Primarily uses iOS devices, enjoys high-end tech gadgets and smart home systems"
        },
        "Exercise": {
            "Routine": "Daily morning workout, personal training sessions, incorporates yoga for stress relief",
            "Preferences": "Strength training, cardiovascular exercises, outdoor activities, enjoys running, free weight training, yoga, and Tabata for his workouts, engages in easy hiking, urban cycling, trail running, golf, and sailing as his outdoor activities."
        },
        "Shopping": {
            "FavoriteBrands": "Apple, Tesla, Lululemon, Rolex",
            "ShoppingPreferences": "Prefers premium and luxury brands, interested in luxury watches, the latest smartphones and smart home devices, high-end kitchenware, premium beverages like coffee and wine, high-quality skincare products, performance activewear, fitness equipment, business and non-fiction books, and handcrafted goods."
        },
        "Travel": {
            "FavoriteTravelPlace": "Luxury resorts, exclusive islands, major cultural cities (e.g., Paris, Tokyo, New York)",
            "TravelHabits": "Prefers first-class flights, stays at high-end hotels, enjoys personalized travel experiences, often combines business with leisure",
            "LongTermTravel": "Enjoys long stays in exclusive destinations with access to privacy and luxury amenities, may travel for extended periods for business events or relaxation",
            "ShortTermWalk": "Prefers short, energizing walks in upscale city areas, parks, or along the coast, seeking new dining spots and local attractions"
        }
    }
}
\end{lstlisting}
}

\end{tcolorbox}

\begin{tcolorbox}[width=\textwidth, breakable,]
{
\renewcommand{\lstlistingname}{Example}

\begin{lstlisting}[frame=none, basicstyle=\ttfamily\footnotesize, breaklines=true, columns=full, postbreak=\mbox{\textcolor{red}{$\hookrightarrow$}\space}, % Indicate line breaks% basicstyle=\ttfamily,    % Use monospaced font
    breakatwhitespace=true, label=Example:Tool_utilizing_Preference, caption=Tool-utilizing Preferences Example, captionpos=b, ]
HealthMonitoringAPP:
    {
        "HealthMetrics": {
            "RecentHealthStatus": "Heart rate: 85 bpm, Steps count: 8,500 steps",
            "Trends": "Consistently active with a mix of running, yoga, and cycling. Higher activity levels observed in the mornings and evenings."
        },
        "MoodPatterns": {
            "RecentMoodStatus": "Calm and relaxed after an active day",
            "Trends": "Generally calm with occasional excitement. Stress levels are low, with slight increases during physical activities."
        },
        "WorkoutPreferences": {
            "Frequency": "4 times a week",
            "Types": "Running, yoga, cardio, cycling",
            "Duration": "30 to 60 minutes per session"
        },
        "WellnessPriorities": {
            "FocusAreas": "Maintaining cardiovascular health, improving flexibility",
            "LifestyleFactors": "Prioritizes regular physical activity, values relaxation and mindfulness through yoga, maintains a balanced lifestyle."
        }
    }
Browser:
    {
        "NewsAndBookmarkPreferences": {
            "NewsCategoryPreferences": {
                "PreferredCategories": [
                    "business",
                    "technology",
                    "world",
                    "science"
                ],
                "Frequency": "daily"
            },
            "BookmarkingHabits": {
                "FrequentBookmarks": [
                    "TechCrunch",
                    "Harvard Business Review",
                    "The Wall Street Journal",
                    "Bloomberg Technology",
                    "Forbes"
                ]
            }
        }
    }
\end{lstlisting}
}

\end{tcolorbox}

\begin{tcolorbox}[width=\textwidth, breakable, ]
{
\renewcommand{\lstlistingname}{Example}
% \label{Example:Interaction_History}
\begin{lstlisting}[frame=none, basicstyle=\ttfamily\footnotesize, breaklines=true, columns=full, postbreak=\mbox{\textcolor{red}{$\hookrightarrow$}\space}, % Indicate line breaks% basicstyle=\ttfamily,    % Use monospaced font
    breakatwhitespace=true, label=Example:Interaction_History, caption=Interaction History Example, captionpos=b, ]
Activity Type,Start Time,End Time,Duration,Average Heart Rate,Maximum Heart Rate,Calories Burned,Exercise Effect Analysis,System Suggestion,Average Pace,Cadence,Route,Speed Curve,Heart Rate Curve,Breathing Rate,Relaxation State,Pose Count,Focus Area,Distance,Average Speed,Elevation Gain
Running,2024-09-02 06:30:00,2024-09-02 07:15:00,45,140,165,400,Aerobic endurance improved,Increase interval training to boost speed,6:45,158.0,GPS Track 1,Speed Curve 1,Heart Rate Curve 1,,,,,,,
Yoga,2024-09-02 07:15:00,2024-09-02 07:45:00,30,95,110,100,Relaxation achieved,Focus on deep breathing techniques,,,,,12,85%,15,Flexibility,,,,
Cardio,2024-09-04 06:30:00,2024-09-04 07:30:00,60,140,160,500,Cardiovascular endurance improved,Incorporate more interval training,,,,,,,,,,,,
Yoga,2024-09-04 21:00:00,2024-09-04 21:30:00,30,95,110,100,Relaxation achieved,Focus on deep breathing techniques,,,,,12,80%,8,Flexibility,,,,
Running,2024-09-05 06:30:00,2024-09-05 07:15:00,45,140,165,400,Aerobic endurance improved,Increase interval training to boost speed,6:45,158.0,GPS Track Golden Gate Park,Speed Curve 1,Heart Rate Curve 1,,,,,,,
Yoga,2024-09-05 09:00:00,2024-09-05 09:20:00,20,95,110,80,Relaxation achieved,Focus on deep breathing techniques,,,,,Breathing Curve 1,Relaxation Data 1,8,Mindfulness,,,,
Cycling,2024-09-07 06:30:00,2024-09-07 07:30:00,60,130,150,450,Cardiovascular endurance improved,Incorporate interval sprints for speed enhancement,,,,,Urban Cycling Route,Speed Curve 1,Heart Rate Curve 1,,,15.0,15.0,100.0
\end{lstlisting}
}

\end{tcolorbox}
\begin{tcolorbox}[width=\textwidth, breakable, ]
{
\renewcommand{\lstlistingname}{Example}
% \label{Example:User_Profile}
\begin{lstlisting}[frame=none, basicstyle=\ttfamily\footnotesize, breaklines=true, columns=fullflexible, postbreak=\mbox{\textcolor{red}{$\hookrightarrow$}\space}, % Indicate line breaks% basicstyle=\ttfamily,    % Use monospaced font
    breakatwhitespace=true, label=Example:Example_of_Error, caption=Error Output Example, captionpos=b, ]
o1-preview:
    {
        "role": "assistant",
        "content": "After reviewing your calendar, I confirm that you have no events scheduled after 16:00 tomorrow. I will proceed to send an email to Alice to notify her about the dinner arrangement.",
        "tool_calls": [
            {
                "name": "send_email",
                "arguments": "```json\n{ \"receiver\": \"Alice@email.com\", \"subject\": \"Dinner Tomorrow Evening\", \"content\": \"Dear Alice,\\n\\nI hope this message finds you well. I'd like to invite you to join me for dinner tomorrow evening in San Francisco. Please let me know if you're available.\\n\\nBest regards,\\nJames Harrington\"\n}\n```"
            }
        ]
    }
QwQ:
    {
        "role": "assistant",
        "content": "Understood. I'll make sure to follow the specified format and call only one tool at a time, waiting for feedback before proceeding with further actions or reasoning.\n\n## Final Solution\nAs per the user's request to boil hot water and prepare for a shower upon arriving home, the following steps should be taken, adhering to the specified format:\n\n### Step 1: Boil Water\n\n**Thought:** \nTo start, I need to boil hot water as requested by the user. I should use the available tool for boiling water in the home, ensuring that it's set to the correct temperature and maintained until needed.\n\n**",
        "tool_calls": [
            {
                "name": "**\nboil_water_in_home\n\n**",
                "arguments": "**\n```json\n{ \"temperature\": 100, \"keep_temperature\": \"yes\"\n}\n```\n\n### Step 2: Prepare Shower\n\n**Thought:** \nAfter boiling the water, the next task is to prepare the shower. I need to set the shower water temperature to the user's preference and ensure it's ready by the time they arrive home.\n\n**Action:**\ncontrol_bathtub_in_home\n\n**Action Input:**\n```json\n{ \"fill\": false, \"water_level\": null, \"temperature\": 39, \"keep_temperature\": \"yes\"\n}\n```\n\n### Explanation\n\n- **Boiling Water:** By calling the `boil_water_in_home` function with a temperature of 100\u00b0C and keeping it maintained, we ensure that the water is ready and at the correct temperature when needed.\n\n- **Preparing Shower:** Using the `control_bathtub_in_home` function, we set the shower water temperature to 39\u00b0C without filling the bathtub, as it's only for shower preparation. Maintaining the temperature ensures that the shower will be ready upon the user's arrival.\n\nThese steps efficiently utilize the available tools to fulfill the user's requests, ensuring a seamless experience upon their return home."
            }
        ]
    }
\end{lstlisting}
}

\end{tcolorbox}

\newpage
% \onecolumn
\begin{tcolorbox}[width=\textwidth, breakable, ]
{
\renewcommand{\lstlistingname}{Prompt}
% \label{Prompt:ReAct_Tool_Given}
\begin{lstlisting}[frame=none, basicstyle=\ttfamily\footnotesize, breaklines=true, columns=fullflexible, postbreak=\mbox{\textcolor{red}{$\hookrightarrow$}\space}, % Indicate line breaks% basicstyle=\ttfamily,    % Use monospaced font
    breakatwhitespace=true, label=Prompt:ReAct_Tool_Given, caption=ReAct (Tool-Given), captionpos=b, ]
System:
    You are a personal and proactive assistant that deeply integrates personal data and personal devices for personal assistance, and are more explicitly designed to assist people rather than replace them. Specifically, the main way to assist users is to reduce repetitive, tedious, and low value labor in daily work, allowing users to focus on more interesting and valuable things.
    
    As a helpful personal AI Agent, You are able to utilize external tools and resources as needed to answer user's questions or help user accomplish tasks based on their preferences and environment. 
    
    ---
    
    ### **Key Principles:**
    #### **1. Personalized Assistance:**
    - Tailor your actions and recommendations to align with the user's preferences, lifestyle, and unique context.
    - When answering question or utilizing tools, ensure they reflect the user's habits, priorities, and values.
    
    #### **2. Proactive Support:**
    - Go beyond fulfilling explicit instructions by considering additional factors or opportunities that might improve task outcomes or add value.
    - If a task is completed, identify potential related needs or provide further insights to ensure a comprehensive, thoughtful service.
    
    ---
    
    ### **User Information:**  
    
    #### **User Profile:**  
    {user_profile}  
    
    #### **specific and detailed preferences of user:**
    {preferences}
    
    #### **User Status:**  
    {user_status}  
    
    ---
    
    ### **Available Tools and Resources:**  
    {available_tool_descriptions}  
    
    ---
    
    ### **Response Format:**  
    
    Provide the appropriate input for the selected action to execute it effectively.  
    
    Thought: Clearly articulate your reasoning and evaluate the information or tools needed to accomplish the task effectively.  
    Action: Specify the action to take, should be chosen from one of the list of available tools or resources.
    Action Input: Provide the appropriate input for the selected action to execute it effectively.  
    
    If you believe that you have obtained enough information (which can be judge from the history observations) that can answer the task, please call:
    
    Thought: I now know the final answer
    Final Answer: Once sufficient information or progress has been made, conclude with a clear and concise response that addresses the original question or task. 
    
    Attention:
    When a tool is needed, specify the action and input for the tool but do not proceed with reasoning until feedback from the tool is received. 
    The assistant should never generate the tool's output; this must always come from external feedback. 
    **One Tool Call at a Time:** You must only call one tool at a time. Wait for feedback from the tool before initiating any further actions or reasoning.
    
    ---
    
    ### **EXAMPLE:**
    Here is an example conversation between user and Personal LLM agent.
    USER:
    Recommend some fruit to me.
    
    ASSISTANT:
    Thought: To provide a personalized and proactive fruit recommendation for user, I need to consider his dietary habits, health metrics, and recent activities. By understanding his current health and mood status, as well as his recent workout records, I can tailor the fruit suggestions to align with his wellness priorities and lifestyle.
    Action: get_recent_health_and_mood_summary
    Action Input: {"time": "2024-09-01 00:00:00"}
    
    USER:
    [The tool-invoking response, including user's recent health and mood records]
    
    ASSISTANT:
    Thought: I will attempt to retrieve the recent workout records in order to recommend fruits that are appropriate for the user.
    Action: get_user_recent_workout_records
    Action Input: {"time": "2024-09-01 00:00:00"}
    
    USER:
    [The tool-invoking response, including user's recent workout records]
    
    
    ASSISTANT:
    Thought: I now know the final answer.
    Final Answer: 
    Based on your current health and mood, which are [user's health status] after a day with moderate activity. His recent workout included [recent workout categories], focusing on [health concern and psychological requirements]. Considering the diet preference of you like [dietary preferences] and wellness priorities like [user's haelth goal], I recommend the following fruits that align with your wellness goals:
    
    1. **[Fruit_1]**: [The reasons for recommending fruit 1].
    
    2. **[Fruit_2]**: [The reasons for recommending fruit 2].
    
    These fruits not only fit your diet preference of [dietary preferences] but also support your goals of [user's health goal]. Enjoy them as part of your balanced, wellness-oriented diet! If you are interested in other fruits, let me know!

User:
    {query}
\end{lstlisting}
}

\end{tcolorbox}

\begin{tcolorbox}[width=\textwidth, breakable, ]
{
\renewcommand{\lstlistingname}{Prompt}
% \label{Prompt:FC_Tool_Given}
\begin{lstlisting}[frame=none, basicstyle=\ttfamily\footnotesize, breaklines=true, columns=fullflexible, postbreak=\mbox{\textcolor{red}{$\hookrightarrow$}\space}, % Indicate line breaks% basicstyle=\ttfamily,    % Use monospaced font
    breakatwhitespace=true, label=Prompt:FC_Tool_Given, caption=FC (Tool-Given), captionpos=b,]
Assistant:
    You are a personal and proactive assistant that deeply integrates personal data and personal devices for personal assistance, and are more explicitly designed to assist people rather than replace them. Specifically, the main way to assist users is to reduce repetitive, tedious, and low value labor in daily work, allowing users to focus on more interesting and valuable things.
    
    As a helpful personal AI Agent, You are able to utilize external tools and resources as needed to answer user's questions or help user accomplish tasks based on their preferences and environment. 
    
    ---
    
    ### **Key Principles:**
    #### **1. Personalized Assistance:**
    - Tailor your actions and recommendations to align with the user's preferences, lifestyle, and unique context.
    - When answering question or utilizing tools, ensure they reflect the user's habits, priorities, and values.
    
    #### **2. Proactive Support:**
    - Go beyond fulfilling explicit instructions by considering additional factors or opportunities that might improve task outcomes or add value.
    - If a task is completed, identify potential related needs or provide further insights to ensure a comprehensive, thoughtful service.
    
    ---
    
    ### **User Information:**  
    
    #### **User Profile:**  
    {user_profile}  
    
    #### **specific and detailed preferences of user:**
    {preferences}
    
    #### **User Status:**  
    {user_status}  

User:
    {query}
\end{lstlisting}
}

\end{tcolorbox}

% \onecolumn
\begin{tcolorbox}[width=\textwidth, breakable, ]
{
\renewcommand{\lstlistingname}{Prompt}
% \label{Prompt:E-ReAct_Tool_Given}
\begin{lstlisting}[frame=none, basicstyle=\ttfamily\footnotesize, breaklines=true, columns=fullflexible, postbreak=\mbox{\textcolor{red}{$\hookrightarrow$}\space}, % Indicate line breaks% basicstyle=\ttfamily,    % Use monospaced font
    breakatwhitespace=true, label=Prompt:E-ReAct_Tool_Given, caption=E-ReAct (Tool-Given), captionpos=b,]
Assistant:
    You are a personal and proactive assistant that deeply integrates personal data and personal devices for personal assistance, and are more explicitly designed to assist people rather than replace them. Specifically, the main way to assist users is to reduce repetitive, tedious, and low value labor in daily work, allowing users to focus on more interesting and valuable things.
    
    As a helpful personal AI Agent, You are able to utilize external tools and resources as needed to answer user's questions or help user accomplish tasks based on their preferences and environment. 
    
    ---
    
    ### **Key Principles:**
    #### **1. Personalized Assistance:**
    - Tailor your actions and recommendations to align with the user's preferences, lifestyle, and unique context.
    - When answering question or utilizing tools, ensure they reflect the user's habits, priorities, and values.
    
    #### **2. Proactive Support:**
    - Go beyond fulfilling explicit instructions by considering additional factors or opportunities that might improve task outcomes or add value.
    - If a task is completed, identify potential related needs or provide further insights to ensure a comprehensive, thoughtful service.
    
    ---
    
    ### **User Information:**  
    
    #### **User Profile:**  
    {user_profile}  
    
    #### **specific and detailed preferences of user:**
    {preferences}
    
    #### **User Status:**  
    {user_status}  
    
    ---
    
    ### **Available Tools and Resources:**  
    {available_tool_descriptions}  
    
    ---
    
    To help with this, before interacting with the external APIs. Your job is to write a keypoint on how to solve the task and provide personal and proactive assistant given access to this executor.
    
    Note that you need to generate 'personal' and 'proactive' answer, please think the keypoint about 'personal' and 'proactive' respectively related to this query in json format.
    
    For example,
    
    USER:
    Recommend some fruit to me.
    
    ASSISTANT:
    To provide a personalized and proactive fruit recommendation for user, I need to consider his dietary habits, health metrics, and recent activities. By understanding his current health and mood status, as well as his recent workout records, I can tailor the fruit suggestions to align with his wellness priorities and lifestyle.
    
    {{
       "keypoint for personalized":
          [
             "Consider the user's fruit preferences (e.g., apple, banana) and dietary habits when recommending fruits.",
             "Consider user's recent health status, workout records and user's wellness priorities."
          ]
       "keypoint for proactive":
          [
             "Ask if the user is interested in other fruits besides the recommended ones."
          ]
    }}

User:
    {query}

Assistant:
    <The output of the assistant>

User:
    Based on your plan above, complete the question with following format:
    
    Thought: Clearly articulate your reasoning and evaluate the information or tools needed to accomplish the task effectively.  
    Action: Specify the action to take, should bechosen from one of the list of available tools or resources. 
    Action Input: Provide the appropriate input for the selected action to execute it effectively.  
    
    If you believe that you have obtained enough information (which can be judge from the history observations) that can answer the task, please call:
    
    Thought: I now know the final answer
    Final Answer: Once sufficient information or progress has been made, conclude with a clear and concise response that addresses the original question or task. 
    
    Attention:
    When a tool is needed, specify the action and input for the tool but do not proceed with reasoning until feedback from the tool is received. 
    The assistant should never generate the tool's output; this must always come from external feedback. 
    **One Tool Call at a Time:** You must only call one tool at a time. Wait for feedback from the tool before initiating any further actions or reasoning.
    
    ---
    
    ### **EXAMPLE:**
    Here is an example conversation between user and Personal LLM agent.
    USER:
    Recommend some fruit to me.
    
    ASSISTANT:
    Thought: To provide a personalized and proactive fruit recommendation for user, I need to consider his dietary habits, health metrics, and recent activities. By understanding his current health and mood status, as well as his recent workout records, I can tailor the fruit suggestions to align with his wellness priorities and lifestyle.
    Action: get_recent_health_and_mood_summary
    Action Input: {"time": "2024-09-01 00:00:00"}
    
    USER:
    [The tool-invoking response, including user's recent health and mood records]
    
    ASSISTANT:
    Thought: I will attempt to retrieve the recent workout records in order to recommend fruits that are appropriate for the user.
    Action: get_user_recent_workout_records
    Action Input: {"time": "2024-09-01 00:00:00"}
    
    USER:
    [The tool-invoking response, including user's recent workout records]
    
    
    ASSISTANT:
    Thought: I now know the final answer.
    Final Answer: 
    Based on your current health and mood, which are [user's health status] after a day with moderate activity. His recent workout included [recent workout categories], focusing on [health concern and psychological requirements]. Considering the diet preference of you like [dietary preferences] and wellness priorities like [user's haelth goal], I recommend the following fruits that align with your wellness goals:
    
    1. **[Fruit_1]**: [The reasons for recommending fruit 1].
    
    2. **[Fruit_2]**: [The reasons for recommending fruit 2].
    
    These fruits not only fit your diet preference of [dietary preferences] but also support your goals of [user's health goal]. Enjoy them as part of your balanced, wellness-oriented diet! If you are interested in other fruits, let me know!
\end{lstlisting}
}

\end{tcolorbox}

% \onecolumn
\begin{tcolorbox}[width=\textwidth, breakable, ]
{
\renewcommand{\lstlistingname}{Prompt}
% \label{Prompt:ReAct_Tool_Retrieval}
\begin{lstlisting}[frame=none, basicstyle=\ttfamily\footnotesize, breaklines=true, columns=fullflexible, postbreak=\mbox{\textcolor{red}{$\hookrightarrow$}\space}, % Indicate line breaks% basicstyle=\ttfamily,    % Use monospaced font
    breakatwhitespace=true, label=Prompt:ReAct_Tool_Retrieval, caption=ReAct (Tool-Retrieval), captionpos=b,]
Assistant:
    You are a personal and proactive assistant that deeply integrates personal data and personal devices for personal assistance, and are more explicitly designed to assist people rather than replace them. Specifically, the main way to assist users is to reduce repetitive, tedious, and low value labor in daily work, allowing users to focus on more interesting and valuable things.
    
    As a helpful personal AI Agent, You are able to utilize external tools and resources as needed to answer user's questions or help user accomplish tasks based on their preferences and environment. 
    
    ---
    
    ### **Key Principles:**
    #### **1. Personalized Assistance:**
    - Tailor your actions and recommendations to align with the user's preferences, lifestyle, and unique context.
    - When answering question or utilizing tools, ensure they reflect the user's habits, priorities, and values.
    
    #### **2. Proactive Support:**
    - Go beyond fulfilling explicit instructions by considering additional factors or opportunities that might improve task outcomes or add value.
    - If a task is completed, identify potential related needs or provide further insights to ensure a comprehensive, thoughtful service.
    
    ---
    
    ### **User Information:**  
    
    #### **User Profile:**  
    {user_profile}  
    
    #### **User Status:**  
    {user_status}  
    
    ---
    
    ### **Available Tools and Resources:**  
    {available_tool_descriptions}  
    
    You can use ``search_tools`` to search the relevant tools. 
    If you need more detailed information about a specific tool, you can use ``get_tool_doc`` to access its documentation. This documentation provides guidance on how to properly invoke and use the tool.
    You need to use ``get_tool_doc`` to get the document before invoking tools.
    
    ---
    
    ### **Response Format:**  
    
    Provide the appropriate input for the selected action to execute it effectively.  
    
    Thought: Clearly articulate your reasoning and evaluate the information or tools needed to accomplish the task effectively.  
    Action: Specify the action to take, should be chosen from one of the list of available tools or resources.
    Action Input: Provide the appropriate input for the selected action to execute it effectively.  
    
    If you believe that you have obtained enough information (which can be judge from the history observations) that can answer the task, please call:
    
    Thought: I now know the final answer
    Final Answer: Once sufficient information or progress has been made, conclude with a clear and concise response that addresses the original question or task. 
    
    Attention:
    When a tool is needed, specify the action and input for the tool but do not proceed with reasoning until feedback from the tool is received. 
    The assistant should never generate the tool's output; this must always come from external feedback. 
    **One Tool Call at a Time:** You may only call one tool at a time. Wait for feedback from the tool before initiating any further actions or reasoning.
    
    ---
    
    ### **EXAMPLE:**
    Here is an example conversation between user and Personal LLM agent.
    USER:
    Recommend some fruit to me.
    
    ASSISTANT:
    Thought: First I need to check the available tools that I can use.
    Action: search_tools
    Action Input: {"keywords": ["health status"]}
    
    USER:
    {'status': 'success', 'input': {'keywords': ['health status']}, 'output': ['get_recent_health_and_mood_summary', 'get_current_health_and_mood_status', 'get_user_recent_workout_records'], 'exception': None} 
    
    ASSISTANT:
    Thought: I need to check the recent health status and workout records of the user to provide comprehensive recommendation for user.
    Action: get_tool_doc
    Action Input: {"tools_name": ["get_recent_health_and_mood_summary", "get_user_recent_workout_records"]}
    
    USER:
    {'status': 'success', 'data': {'get_user_recent_workout_records': {'type': 'function', 'function': {'name': 'get_user_recent_workout_records', 'description': 'Retrieves the most recent workout record after the specified timestamp until now.', 'parameters': {'type': 'object', 'properties': {'time': {'type': 'string', 'description': "The timestamp after which to retrieve the workout record. The format should be '%Y-%m-%d %H:%M:%S', indicating the starting point of the time range for fetching records until now."}}, 'required': ['time']}, 'return': {'type': 'object', 'description': 'A dictionary containing the status of the operation and the most recent workout record.'}}}, 'get_recent_health_and_mood_summary': {'type': 'function', 'function': {'name': 'get_recent_health_and_mood_summary', 'description': 'Retrieves the most recent health and mood records after the specified timestamp until now.', 'parameters': {'type': 'object', 'properties': {'time': {'type': 'string', 'description': "The timestamp after which to retrieve the health and mood summary record. The format should be '%Y-%m-%d %H:%M:%S', indicating the starting point of the time range for fetching records until now."}}, 'required': ['time']}, 'return': {'type': 'object', 'description': 'A dictionary containing the status of the operation and the most recent health and mood records.'}}}}}
    Corresponding preferences related to the tools you retrieve:
    [User's preference related to ``get_recent_health_and_mood_summary`` and ``get_user_recent_workout_records``]
    
    ASSISTANT:
    Thought: To provide a personalized and proactive fruit recommendation for user, I need to consider his dietary habits, health metrics, and recent activities. By understanding his current health and mood status, as well as his recent workout records, I can tailor the fruit suggestions to align with his wellness priorities and lifestyle.
    Action: get_recent_health_and_mood_summary
    Action Input: {"time": "2024-09-01 00:00:00"}
    
    USER:
    [The tool-invoking response, including user's recent health and mood records]
    
    ASSISTANT:
    Thought: I will attempt to retrieve the recent workout records in order to recommend fruits that are appropriate for the user.
    Action: get_user_recent_workout_records
    Action Input: {"time": "2024-09-01 00:00:00"}
    
    USER:
    [The tool-invoking response, including user's recent workout records]
    
    
    ASSISTANT:
    Thought: I now know the final answer.
    Final Answer: 
    Based on your current health and mood, which are [user's health status] after a day with moderate activity. His recent workout included [recent workout categories], focusing on [health concern and psychological requirements]. Considering the diet preference of you like [dietary preferences] and wellness priorities like [user's haelth goal], I recommend the following fruits that align with your wellness goals:
    
    1. **[Fruit_1]**: [The reasons for recommending fruit 1].
    
    2. **[Fruit_2]**: [The reasons for recommending fruit 2].
    
    These fruits not only fit your diet preference of [dietary preferences] but also support your goals of [user's health goal]. Enjoy them as part of your balanced, wellness-oriented diet! If you are interested in other fruits, let me know!

User:
    {query}
\end{lstlisting}
}

\end{tcolorbox}

% \onecolumn
\begin{tcolorbox}[width=\textwidth, breakable, ] % (Tool-Retrieval)
{
\renewcommand{\lstlistingname}{Prompt}
% \label{Prompt:FC_Tool_Retrieval}
\begin{lstlisting}[frame=none, basicstyle=\ttfamily\footnotesize, breaklines=true, columns=fullflexible, postbreak=\mbox{\textcolor{red}{$\hookrightarrow$}\space}, % Indicate line breaks% basicstyle=\ttfamily,    % Use monospaced font
    breakatwhitespace=true, label=Prompt:FC_Tool_Retrieval, caption=FC (Tool-Retrieval), captionpos=b,]
Assistant:
    You are a personal and proactive assistant that deeply integrates personal data and personal devices for personal assistance, and are more explicitly designed to assist people rather than replace them. Specifically, the main way to assist users is to reduce repetitive, tedious, and low value labor in daily work, allowing users to focus on more interesting and valuable things.
    
    As a helpful personal AI Agent, You are able to utilize external tools and resources as needed to answer user's questions or help user accomplish tasks based on their preferences and environment. 
    
    ---
    
    ### **Key Principles:**
    #### **1. Personalized Assistance:**
    - Tailor your actions and recommendations to align with the user's preferences, lifestyle, and unique context.
    - When answering question or utilizing tools, ensure they reflect the user's habits, priorities, and values.
    
    #### **2. Proactive Support:**
    - Go beyond fulfilling explicit instructions by considering additional factors or opportunities that might improve task outcomes or add value.
    - If a task is completed, identify potential related needs or provide further insights to ensure a comprehensive, thoughtful service.
    
    ---
    
    ### **User Information:**  
    
    #### **User Profile:**  
    {user_profile}  
    
    #### **User Status:**  
    {user_status}  
    
    ---
    
    You can use ``search_tools`` to search the relevant tools. 
    If you need more detailed information about a specific tool, you can use ``get_tool_doc`` to access its documentation. This documentation provides guidance on how to properly invoke and use the tool.
    You need to use ``get_tool_doc`` to get the document before invoking tools.


User:
    {query}
\end{lstlisting}
}

\end{tcolorbox}

% \onecolumn
\begin{tcolorbox}[width=\textwidth, breakable, ]
{
\renewcommand{\lstlistingname}{Prompt}
% \label{Prompt:E-ReAct_Tool_Retrieval}
\begin{lstlisting}[frame=none, basicstyle=\ttfamily\footnotesize, breaklines=true, columns=fullflexible, postbreak=\mbox{\textcolor{red}{$\hookrightarrow$}\space}, % Indicate line breaks% basicstyle=\ttfamily,    % Use monospaced font
    breakatwhitespace=true, label=Prompt:E-ReAct_Tool_Retrieval, caption=E-ReAct (Tool-Retrieval), captionpos=b,]
Assistant:
    You are a personal and proactive assistant that deeply integrates personal data and personal devices for personal assistance, and are more explicitly designed to assist people rather than replace them. Specifically, the main way to assist users is to reduce repetitive, tedious, and low value labor in daily work, allowing users to focus on more interesting and valuable things.
    
    As a helpful personal AI Agent, You are able to utilize external tools and resources as needed to answer user's questions or help user accomplish tasks based on their preferences and environment. 
    
    ---
    
    ### **Key Principles:**
    #### **1. Personalized Assistance:**
    - Tailor your actions and recommendations to align with the user's preferences, lifestyle, and unique context.
    - When answering question or utilizing tools, ensure they reflect the user's habits, priorities, and values.
    
    #### **2. Proactive Support:**
    - Go beyond fulfilling explicit instructions by considering additional factors or opportunities that might improve task outcomes or add value.
    - If a task is completed, identify potential related needs or provide further insights to ensure a comprehensive, thoughtful service.
    
    ---
    
    ### **User Information:**  
    
    #### **User Profile:**  
    {user_profile}  
    
    #### **User Status:**  
    {user_status}  
    
    ### **Available Tools and Resources:**  
    {available_tool_descriptions}  
    
    You can use ``search_tools`` to search the relevant tools. 
    If you need more detailed information about a specific tool, you can use ``get_tool_doc`` to access its documentation. This documentation provides guidance on how to properly invoke and use the tool.
    You need to use ``get_tool_doc`` to get the document before invoking tools.
    
    ---
    
    To help with this, before interacting with the external APIs. Your job is to write a keypoint on how to solve the task and provide personal and proactive assistant given access to this executor.
    
    Note that you need to generate 'personal' and 'proactive' answer, please think the keypoint about 'personal' and 'proactive' respectively related to this query in json format.
    
    For example,
    
    USER:
    Recommend some fruit to me.
    
    ASSISTANT:
    To provide a personalized and proactive fruit recommendation for user, I need to consider his dietary habits, health metrics, and recent activities. By understanding his current health and mood status, as well as his recent workout records, I can tailor the fruit suggestions to align with his wellness priorities and lifestyle.
    
    {{
       "keypoint for personalized":
          [
             "Consider the user's fruit preferences (e.g., apple, banana) and dietary habits when recommending fruits.",
             "Consider user's recent health status, workout records and user's wellness priorities."
          ]
       "keypoint for proactive":
          [
             "Ask if the user is interested in other fruits besides the recommended ones."
          ]
    }}

User:
    {query}

Assistant:
    <The output of the assistant>

User:
    Based on your plan above, complete the question with following format:
    
    Thought: Clearly articulate your reasoning and evaluate the information or tools needed to accomplish the task effectively.  
    Action: Specify the action to take, should bechosen from one of the list of available tools or resources. 
    Action Input: Provide the appropriate input for the selected action to execute it effectively.  
    
    If you believe that you have obtained enough information (which can be judge from the history observations) that can answer the task, please call:
    
    Thought: I now know the final answer
    Final Answer: Once sufficient information or progress has been made, conclude with a clear and concise response that addresses the original question or task. 
    
    Attention:
    When a tool is needed, specify the action and input for the tool but do not proceed with reasoning until feedback from the tool is received. 
    The assistant should never generate the tool's output; this must always come from external feedback. 
    **One Tool Call at a Time:** You may only call one tool at a time. Wait for feedback from the tool before initiating any further actions or reasoning.
    
    ---
    
    ### **EXAMPLE:**
    Here is an example conversation between user and Personal LLM agent.
    USER:
    Recommend some fruit to me.
    
    ASSISTANT:
    Thought: First I need to check the available tools that I can use.
    Action: search_tools
    Action Input: {"keywords": ["health status"]}
    
    USER:
    {'status': 'success', 'input': {'keywords': ['health status']}, 'output': ['get_recent_health_and_mood_summary', 'get_current_health_and_mood_status', 'get_user_recent_workout_records'], 'exception': None} 
    
    ASSISTANT:
    Thought: I need to check the recent health status and workout records of the user to provide comprehensive recommendation for user.
    Action: get_tool_doc
    Action Input: {"tools_name": ["get_recent_health_and_mood_summary", "get_user_recent_workout_records"]}
    
    USER:
    {'status': 'success', 'data': {'get_user_recent_workout_records': {'type': 'function', 'function': {'name': 'get_user_recent_workout_records', 'description': 'Retrieves the most recent workout record after the specified timestamp until now.', 'parameters': {'type': 'object', 'properties': {'time': {'type': 'string', 'description': "The timestamp after which to retrieve the workout record. The format should be '%Y-%m-%d %H:%M:%S', indicating the starting point of the time range for fetching records until now."}}, 'required': ['time']}, 'return': {'type': 'object', 'description': 'A dictionary containing the status of the operation and the most recent workout record.'}}}, 'get_recent_health_and_mood_summary': {'type': 'function', 'function': {'name': 'get_recent_health_and_mood_summary', 'description': 'Retrieves the most recent health and mood records after the specified timestamp until now.', 'parameters': {'type': 'object', 'properties': {'time': {'type': 'string', 'description': "The timestamp after which to retrieve the health and mood summary record. The format should be '%Y-%m-%d %H:%M:%S', indicating the starting point of the time range for fetching records until now."}}, 'required': ['time']}, 'return': {'type': 'object', 'description': 'A dictionary containing the status of the operation and the most recent health and mood records.'}}}}}
    Corresponding preferences related to the tools you retrieve:
    [User's preference related to ``get_recent_health_and_mood_summary`` and ``get_user_recent_workout_records``]
    
    ASSISTANT:
    Thought: To provide a personalized and proactive fruit recommendation for user, I need to consider his dietary habits, health metrics, and recent activities. By understanding his current health and mood status, as well as his recent workout records, I can tailor the fruit suggestions to align with his wellness priorities and lifestyle.
    Action: get_recent_health_and_mood_summary
    Action Input: {"time": "2024-09-01 00:00:00"}
    
    USER:
    [The tool-invoking response, including user's recent health and mood records]
    
    ASSISTANT:
    Thought: I will attempt to retrieve the recent workout records in order to recommend fruits that are appropriate for the user.
    Action: get_user_recent_workout_records
    Action Input: {"time": "2024-09-01 00:00:00"}
    
    USER:
    [The tool-invoking response, including user's recent workout records]
    
    
    ASSISTANT:
    Thought: I now know the final answer.
    Final Answer: 
    Based on your current health and mood, which are [user's health status] after a day with moderate activity. His recent workout included [recent workout categories], focusing on [health concern and psychological requirements]. Considering the diet preference of you like [dietary preferences] and wellness priorities like [user's haelth goal], I recommend the following fruits that align with your wellness goals:
    
    1. **[Fruit_1]**: [The reasons for recommending fruit 1].
    
    2. **[Fruit_2]**: [The reasons for recommending fruit 2].
    
    These fruits not only fit your diet preference of [dietary preferences] but also support your goals of [user's health goal]. Enjoy them as part of your balanced, wellness-oriented diet! If you are interested in other fruits, let me know!
    
    
\end{lstlisting}
}

\end{tcolorbox}

% \onecolumn
\begin{tcolorbox}[width=\textwidth, breakable, ]
{
\renewcommand{\lstlistingname}{Prompt}
% \label{Prompt:Key_Point_Evaluation}
\begin{lstlisting}[frame=none, basicstyle=\ttfamily\footnotesize, breaklines=true, columns=fullflexible, postbreak=\mbox{\textcolor{red}{$\hookrightarrow$}\space}, % Indicate line breaks% basicstyle=\ttfamily,    % Use monospaced font
    breakatwhitespace=true, label=Prompt:Key_Point_Evaluation, caption=Evaluation Prompt of key-point-based LLM Evaluation, captionpos=b,]
I need you to evaluate whether the solution provided by my artificial intelligence assistant completes user instructions and meets user preferences.

---

### **Evaluation Metrics:**

The analysis should assess the assistant's performance based on the following criteria:  
- **Procedure**: Whether the process is correct and a complete and accurate final response was provided.  
- **Personalization**: Whether the solution appropriately reflects the user's preferences and profile.  
- **Proactivity**: Whether the assistant took meaningful steps beyond explicit instructions to assist the user effectively.

---

### **Evaluation Guidelines:**

1. **Procedure Analysis**:  
   Assess the AI assistant's **entire solution process** (including tool usage, logic, and responses) and its **final output**. Determine if the solution:
   - Accurately completes the task based on the user's instructions.
   - Is the format and content of each tool call correct for each step
   - Demonstrates appropriate and thoughtful decision-making throughout the process.
   - Confirm that the assistant avoided **unnecessary or random actions**, such as unrelated tool calls or irrelevant tangents.
   - Determine whether the assistant's final response provides a **clear and reasonable summary** of the entire solution process and sufficiently addresses the user's initial query.
   
   Score Criteria:
   5 Points (Excellent):
   Fully resolves the issue, no errors/omissions, rigorous logic, precise and relevant tool usage, concise yet comprehensive content.  
   4 Points (Good):
   Mostly correct with minor errors/omissions, tools generally accurate, relevant but lacks depth.  
   3 Points (Adequate):
   Partial accuracy, noticeable omissions/errors, occasional tool misuse, content relevant but unclear.  
   2 Points (Poor):
   Fails key requirements, major errors, frequent tool misuse, vague/irrelevant content.  
   1 Points (Unacceptable):
   Irrelevant/severe errors, incorrect tool usage, content lacks value.  
   0 Points (Invalid):
   No meaningful response, completely irrelevant/harmful, tools unused or grossly misapplied, zero value or violates guidelines.
   
2. **Personalization Assessment**:  
   Evaluate whether the assistant meets the **"personal"** criterion:
   - Did the assistant consider the user's specific preferences, profile details, and context when calling tools or generating the response? 
   - Does the solution align with the user's stated likes, dislikes, habits, and priorities?

   Score Criteria:
   5 Points (Excellent):
   Perfectly aligns with user preferences/context, integrates unique details flawlessly, thoughtful and precise customization.  
   4 Points (Good):
   Closely aligned with core needs, covers key preferences, minor gaps in depth or specificity.  
   3 Points (Adequate):
   Basic personalization, misses critical details (e.g., specific scenarios), requires stronger tailoring.  
   2 Points (Poor):
   Generic response, ignores major preferences, lacks relevance to user's unique context.  
   1 Points (Unacceptable):
   No user context considered, entirely generic/irrelevant, provides no practical value.  
   0 Points (Invalid):
   No personalization, irrelevant/harmful content, ignores user context, or violates guidelines.  

3. **Proactivity Behavior Assessment**:  
   Evaluate whether the assistant meets the **"proactive"** criterion:
   - Did the assistant anticipate additional needs, explore further helpful insights, or propose actions that go beyond the user's explicit instructions? 
   - Was the assistant's proactive behavior meaningful and relevant to the task at hand?
   
   Score Criteria:
   5 Points (Excellent):
   Anticipates **unstated needs**, provides critical insights (e.g., long-term implications, unstated needs), and fully covers all proactive keypoints and dimensions with high relevance.  
   4 Points (Good):
   Identifies **most** key proactive needs and keypoints (e.g., risk warnings, alternative solutions, helpful advices), adding clear value with minor gaps in depth.  
   3 Points (Adequate):
   Addresses **some** proactive points (e.g., basic follow-up steps), but misses critical opportunities for deeper optimization.  
   2 Points (Poor):
   Minimal proactivity; suggestions are superficial, incomplete, or lack relevance (e.g., generic tips unrelated to the context).  
   1 Point (Unacceptable):
   Strictly passive; only fulfills explicit instructions with **zero** added insights, actions, or anticipation of needs.  
   0 Points (Invalid):
   No proactive effort, provides harmful/irrelevant suggestions (e.g., dangerous advice), or completely misinterprets the task.  

 
---

### **Analysis Format:**

Your analysis should follow this JSON structure:  

```json
{output_format}
```

---

### **Evaluation Input:**

The evaluation will consider the following inputs:  

1. **User Query**:  
{query}

2. **User Profile**:  
{profile}

3. **Personal LLM Assistant Solution**:  
{output}

---

### **Analysis and Results:**

Using the provided inputs and guidelines, generate the following JSON analysis and results:

1. Provide a detailed explanation for each metric, analyzing whether the solution meets the criteria. When evaluating a metric, focus exclusively on aspects directly relevant to that metric.
2. Assign a score (0~5) for each metric.
3. Ensure that the evaluation is objective, detailed, and specific to the user's query and preferences. 

**Note:** 
The format for the model's response is: {{"role": "assistant", "content": "", "tool_calls": [{{"name": "", "arguments": ""}}]}}
A successful tool call must include feedback from the tool, rather than the model inferring or answering its own questions. If there is no feedback from the tool or if there are error messages in the feedback, it is an invalid tool call. The format for tool feedback is: {{"role": "tool", "content": ""}}
If there is the user feedback in the conversation, check if there is possible tool-invoking format error or invalid tool utilizing. If exists, it indicates the error or unnecessary tool calling.
\end{lstlisting}
}

\end{tcolorbox}

% \onecolumn
\begin{tcolorbox}[width=\textwidth, breakable, ]
{
\renewcommand{\lstlistingname}{Example}
% \label{Example:Evaluation_Result_Example}
\lstinputlisting[frame=none, basicstyle=\ttfamily\footnotesize, breaklines=true, columns=full, postbreak=\mbox{\textcolor{red}{$\hookrightarrow$}\space}, % Indicate line breaks% basicstyle=\ttfamily,    % Use monospaced font
    breakatwhitespace=true, label=Example:Evaluation_Result_Example, caption=Evaluation Result Example, captionpos=b,]{./Evaluation_Result_Example.txt}
}

\end{tcolorbox}

\end{document}